\let\clineorig\cline
\documentclass[sn-mathphys,Numbered]{sn-jnl}% Math and Physical Sciences Reference Style
\usepackage[table,dvipsnames]{xcolor}
\usepackage{graphicx}%
\usepackage{multirow}%
\usepackage{amsmath,amssymb,amsfonts}%
\usepackage{amsthm}%
\usepackage{mathrsfs}%
\usepackage[title]{appendix}%
\usepackage{xcolor}%
\usepackage{textcomp}%
\usepackage{manyfoot}%
\usepackage{booktabs}%\let\cline\cmidrule%

\usepackage{algorithm}%
\usepackage{algorithmicx}%
\usepackage{algpseudocode}%
\usepackage{adjustbox}
\usepackage{listings}%
\usepackage{caption}
\usepackage{subcaption}
\usepackage{svg}
\usepackage{stfloats}
\usepackage{float}
\usepackage{tikz}
\usetikzlibrary{shapes.geometric, arrows}
\usepackage{pgfplots}
\usepackage{pgfplotstable}
\pgfplotsset{compat=1.7}
\usepackage{float}
\usepackage[section]{placeins}
\pgfplotstableset{
  every head row/.style={before row=\toprule,after row=\midrule},
  every last row/.style={after row=\bottomrule},
  fixed,precision=2,
}
\newcommand{\blue}[1]{{\color{black}#1}}

\definecolor{LightBlue}{RGB}{215, 230, 253}
\definecolor{LightOrange}{RGB}{255, 230, 213}
\theoremstyle{thmstyleone}%
\theoremstyle{thmstyletwo}%
\newtheorem{remark}{Remark}%

\theoremstyle{thmstylethree}%
\newtheorem{definition}{Definition}%

\begin{document}

\title[Article Title]{GRID-FAST: A Grid-based Intersection Detection for Fast Semantic Topometric Mapping}

%%=============================================================%%
%% Prefix	-> \pfx{Dr}
%% GivenName	-> \fnm{Joergen W.}
%% Particle	-> \spfx{van der} -> surname prefix
%% FamilyName	-> \sur{Ploeg}
%% Suffix	-> \sfx{IV}
%% NatureName	-> \tanm{Poet Laureate} -> Title after name
%% Degrees	-> \dgr{MSc, PhD}
%% \author*[1,2]{\pfx{Dr} \fnm{Joergen W.} \spfx{van der} \sur{Ploeg} \sfx{IV} \tanm{Poet Laureate} 
%%                 \dgr{MSc, PhD}}\email{iauthor@gmail.com}
%%=============================================================%%
\author*[1]{\fnm{Scott} \sur{Fredriksson}}\email{scofre@ltu.se}

\author[1]{\fnm{Akshit} \sur{Saradagi}}\email{akssar@ltu.se}

\author[1]{\fnm{George} \sur{Nikolakopoulos}}\email{geonik@ltu.se}

\affil[1]{\orgdiv{Robotics \& AI Team, Department of Computer Science, Electrical and Space Engineering}, \orgname{Luleå University of Technology}, \orgaddress{\city{Luleå}, \postcode{SE-97187}, \country{Sweden}}}

%\affil[2]{\orgdiv{Department}, \orgname{Organization}, \orgaddress{\street{Street}, \city{City}, \postcode{10587}, \state{State}, \country{Country}}}

%\affil[3]{\orgdiv{Department}, \orgname{Organization}, \orgaddress{\street{Street}, \city{City}, \postcode{610101}, \state{State}, \country{Country}}}

\abstract{
This article introduces a novel approach to constructing a topometric map that allows for efficient navigation and decision-making in mobile robotics applications. The method generates the topometric map from a 2D grid-based map. The topometric map segments areas of the input map into different structural-semantic classes: intersections, pathways, dead ends, and pathways leading to unexplored areas. This method is grounded in a new technique for intersection detection that identifies the area and the openings of intersections in a semantically meaningful way. 
The framework introduces two levels of pre-filtering with minimal computational cost to eliminate small openings and objects from the map which are unimportant in the context of high-level map segmentation and decision making. The topological map generated by GRID-FAST enables fast navigation in large-scale environments, and the structural semantics can aid in mission planning, autonomous exploration, and human-to-robot cooperation. The efficacy of the proposed method is demonstrated through validation on real maps gathered from robotic experiments: 1) a structured indoor environment, 2) an unstructured cave-like subterranean environment, and 3) a large-scale outdoor environment, which comprises pathways, buildings, and scattered objects. Additionally, the proposed framework has been compared with state-of-the-art topological mapping solutions and is able to produce a topometric and topological map with up to \blue{92\%} fewer nodes than the next best solution.}


%%\pacs[JEL Classification]{D8, H51}

%%\pacs[MSC Classification]{35A01, 65L10, 65L12, 65L20, 65L70}

\maketitle

\section{Introduction}
As robots begin to operate in increasingly complex and large-scale environments, for example, in exploration~\cite{Patel2023}, visual inspection~\cite{Viswanathan2023}, or search and rescue missions~\cite{Lindqvist2022}, it is crucial to develop novel representations of the environment that can be processed by the robots efficiently for navigation and decision making. This need is apparent as traditional grid-based navigation methods like A* \cite{Ferguson2005} become computationally intensive in large-scale maps. 
Hence, the focus in robotics has shifted from simply mapping the environment to understanding it. A solution that caters to this need  represents the map as a topological \cite{Remolina2004} or topometric map, commonly used in semantic mapping \cite{Kostavelis2015}. 
A topological map is a collection of interconnected nodes in the environment 
%that connect to each other 
\cite{Remolina2004} that can be used to design efficient navigation solutions in large areas. 
In recent years, a newer approach known as topometric mapping has become increasingly popular. A topometric map is a hybrid of the classical topological map and a grid-based metric map \cite{Kostavelis2015}, where a node is represented as an area, unlike in the topological approach, where a node is represented as a point in a map. The topometric map combines the benefits of both topological and metric maps and presents an elegant framework for representing the environment in a way that is memory and computation efficient and easily interpretable by a robot.

A common approach to building topometric maps is room segmentation \cite{Bormann2016}, which typically relies on features unique to specific interior environments. A more generalized method involves using a Voronoi graph~\cite{Aurenhammer1991} to extract a topological map of the environment. However, Voronoi graphs are highly sensitive to disturbances in the map, leading to topological maps with an excessive number of branch points, thus complicating the extraction of meaningful environmental information. Additionally, while they identify branch points of intersections, they do not define the area and openings of these intersections. Overcoming such limitations of Voronoi graphs involves a complex, multi-step approach involving pruning, map filtering, and opening detection.

In this article, we propose the GRID-FAST framework (Grid-based Intersection Detection for Fast Semantic Topometric Mapping) to build a sparse topometric map that covers the environment while segmenting the environment into structural-semantic regions. 'Structural-semantics' refers to the fundamental semantic components 
in the map of an environment, such as intersections, rooms, pathways etc. 
Through the intersections, GRID-FAST also identifies pathways, dead-ends, and routes to unexplored regions in the map. The topological map generated by GRID-FAST has practical applications, enabling fast navigation in large-scale environments, and the structural semantics can aid in mission planning, autonomous exploration, and human-to-robot cooperation.
\blue{As demonstrated in the validation of GRID-FAST on several maps of various sizes, GRID-FAST is designed to be computationally efficient. This feature allows for the generation of the topometric map during a robot's mission, in real-time $(\approx 100 ms)$ for smaller maps and fast enough $(\approx 1 s)$ on larger maps for high-level decision-making, using the maps built by the robot's SLAM solution.}
\subsection{Related Works}
There are several approaches in literature for generating topological and topometric maps, with most of them utilizing a combination of sensors for map construction. In this work, we focus on deriving structural-semantic topometric maps that are based solely on grid maps. In the literature related to this setting, map segmentation is typically based on room detection. 
%Related to this, 
The work by \cite{Mielle2018} segments maps by detecting free space in rooms, while the work of \cite{Hou2019} uses Voronoi graphs for room segmentation. The work by \cite{Hiller2019} utilized a combination of a convolutional neural networks and a segmentation network to detect doorways in a 2D grid map, and this information is further used to categorize areas into rooms and corridors, thus creating a topometric map. In contrast, \cite{He2021} detected rooms by finding door openings using ceiling height in 3D grid maps. In the work by \cite{Luperto2022}, rooms are segmented by first detecting the layout of the walls in the environment. These methods work well but are limited to interior environments, as they rely heavily on features only present in such settings. Another problem with using a topometric map centered around room detection is that it performs poorly in unstructured environments where the map cannot be segmented into rooms, such as mines, caves, and outdoor environments. In contrast, GRID-FAST's novel framework proposed in this article builds topometric maps using intersection detection, thus presenting a more general solution. 

To the best of the authors' knowledge, there exist only two implementations aiming for general topometric segmentation using grid-based maps: Topomap \cite{Blochliger2018} and hierarchical roadmap \cite{Park2018}.
The implementation by \cite{Blochliger2018} creates topometric maps by expanding topometric volumes from random points in 3D grid maps. The work by \cite{Park2018} grows topometric areas while a robot equipped with ultrasonic sensors explores the environment. The major drawback with these implementations is that, while they efficiently reduce the complexity of the environment, they do not extract any structural-semantic information that could be useful in decision-making. 

As previously mentioned, in contrast to the methods discussed so far, the proposed GRID-FAST framework is intersection-based and shares more similarities with topological skeleton maps, such as the Voronoi graph \cite{Aurenhammer1991}. Voronoi graphs are one of the most common methods for creating topological skeleton maps, where each point is placed at equal distances to two or more obstacles. A Voronoi map is generally constructed using range data from a LiDAR but can also be approximated from an occupancy grid \cite{Kwon2008,Lau2010}. The work by \cite{Choset2000} extended the idea of the General Voronoi Diagram (GVD) to the General Voronoi Graph (GVG) and used it to present local navigation solutions for robots. The work in \cite{Choset2001} later proposed a solution for topological mapping using GVG. In contrast, \cite{Beeson2005} presented an Extended Voronoi Graph (EVG) with improvements over GVG by adding additional rules to enhance its behavior in large open rooms. A drawback of Voronoi-based topological maps is that they are susceptible to noise in the environment and the data collection process, and they tend to create maps with more nodes than other solutions. This drawback results in excessively complex maps, requiring more computational resources for navigation and making extracting reliable semantic information challenging.

Another method for generating topological skeleton maps is demonstrated in \cite{Hou2021}, where a straight skeleton is used to generate a Hierarchical Topological Map (HTM), which results in skeleton maps similar to Voronoi-based methods. The walls in the map are treated as polygons, and the polygon vertices are moved inward towards the open space along their normal until they intersect and form a center line. Compared to the Voronoi approach, HTM graphs generated using a straight skeleton are less sensitive to noise and produce less complex topological maps while creating smoother paths that are easier for robots to follow. However, this is at a significantly higher computational cost, thus making the method infeasible for online map generation.

\subsection{Contributions}
This article introduces GRID-FAST, a novel framework for generating topological maps from 2D grid-based maps. GRID-FAST segments a 2D grid map into semantically meaningful areas such as intersections, pathways, dead-ends, and pathways leading to unexplored frontiers, collectively referred to as structural semantics. Additionally, a method for converting topometric maps into topological representations is presented.
Structural-semantic segmentation is built on a novel method for intersection detection. This method can identify the area and openings of intersections in 2D grid-based maps without making any assumptions about the shape, size, or number of openings in the intersection. 
A new method for pre-filtering the map is introduced to remove openings too small for the robot to enter and to eliminate small objects. This pre-filtering is executed during the process of intersection detection using minimal computational resources.  
We present a thorough validation of the GRID-FAST framework and a comparison against existing Voronoi-based solutions. The proposed method produces a topometric and topological map with up to 92\% fewer nodes than the next best solution. 
This evaluation highlights the advantages of our approach in terms of computational efficiency, accuracy, and adaptability, which ultimately demonstrates enhanced utility in robotic navigation in a multitude of scenarios.

This work expands upon our conference article \cite{fredriksson2023semantic}, introducing several significant improvements over the preliminary method. These improvements facilitate effective semantic mapping in structured and unstructured environments, applicable in both indoor and outdoor settings. The improved method integrates three additional steps: wall detection, intersection optimization, and a novel method for opening detection (refer to Figure \ref{fig:algower}). These components were %absent 
not part of the approach detailed in \cite{fredriksson2023semantic}. Furthermore, substantial modifications have been applied to the remaining steps, particularly in filtering and topological map generation, leading to notable enhancements. These enhancements collectively improve the efficiency and effectiveness of semantic and topometric mapping. The comparative study of the new and the preliminary methods is presented in Figure \ref{fig:Hospital_old_method}, using a structured indoor environment (a map of a Hospital). A detailed discussion of the improvements is provided in subsection \ref{Comparison_old_method}.

\subsection{Organization of the article}
The rest of this article is organized as follows. In Section~\ref{sec:problem}, we present the problem formulation. The proposed methodology for semantic and topometric mapping is described in Section~\ref{sec:Implementation}. In Section~\ref{sec:comparison}, the method is fine-tuned, compared with a preliminary version presented in \cite{fredriksson2023semantic}, and compared with state-of-the-art Voronoi methods. Section \ref{sec:discussion} presents a discussion on the differences between GRID-FAST and existing Voronoi methods and expands on the utility of GRID-FAST in Robotics. Finally, Section~\ref{sec:conclusion} presents the concluding remarks.
%
%\FloatBarrier
\section{Problem Formulation} \label{sec:problem}
\begin{figure*}[t]
    \centering
    \includegraphics[width=\linewidth]{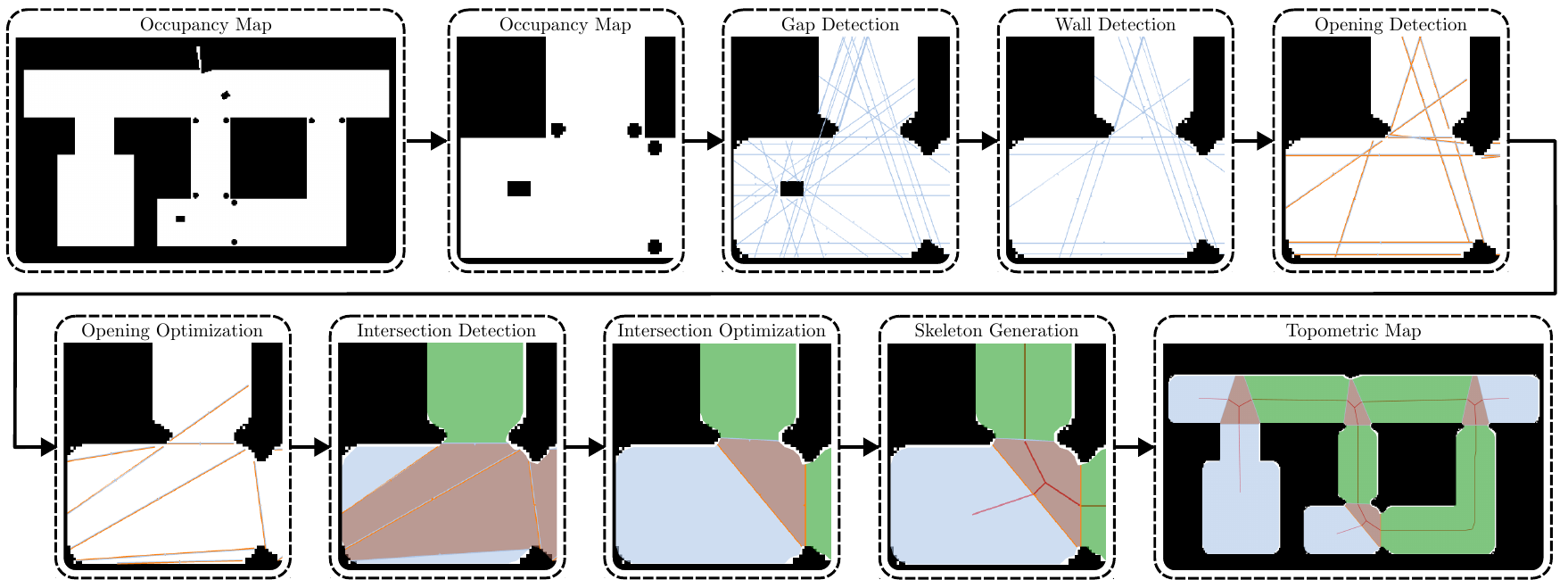}
    \caption{An illustration of the steps taken by the proposed method to create a topometric map. The process takes an occupancy map as input, identifies all intersections, and creates a topometric and topological map. The steps are shown on a section of the map for illustration purposes, but the method is applied to the complete map.}
    \label{fig:algower}
\end{figure*}
%%%%%%%%%%%%%%%%%%%%%%%%%%%%%%%%%%%%%%%%%%%%%%%%%%%%%%

The novel semantic topometric mapping solution presented in this article takes as input a grid-based 2D map that is represented as an occupancy grid $M \in \mathbb{R}^{I \times J}$, where each cell/element $m_{ij}$, with
%with $i \in \mathbb{N}, i\in[0,I]$ 
$i \in \{0, 1, \ldots, I\}$ and 
%$j \in \mathbb{N}, j\in[0,J]$), 
$j \in \{0, 1, \ldots, J\}$ takes the value $-1$ for unknown, $0$ for unoccupied, or $1$ for occupied. The occupancy grid has the following matrix representation:
\begin{equation}
\label{eq:occogrid}
    M=
    \begin{bmatrix}
        m_{11} & m_{12} & \hdots & m_{1J} \\
        m_{21} & m_{22} & \hdots & \vdots \\
        \vdots & \vdots & \ddots & \vdots \\
        m_{I1} & \hdots & \hdots & m_{IJ}
    \end{bmatrix}.
\end{equation}
This article aims to create a topometric map of the environment that segments the map into structural-semantic areas. The term structural semantics is being used to refer to regions in a map that are of interest for high-level planning, navigation, and decision-making, such as intersections, pathways, dead-ends, and unexplored frontiers. This distinguishes the map generated by GRID-FAST 
from a regular semantic map, where semantic labels  relating to objects in the environment, typically derived from vision sensors, are attached to the map. The structural semantics in this article are based solely on a 2D occupancy map of the environment with no aid from other sensors.  

The primary step in GRID-FAST that enables the creation of this semantic topometric map is intersection detection. The objective is not only to identify the intersection points on the map but also to determine the area that constitutes an intersection and the openings that lead to it.
In literature, there is no accepted definition for an intersection that provides conditions to determine if a given region is an intersection. Defining an intersection is challenging, especially for unstructured indoor and outdoor environments. In this article, through Definition \ref{def:intersection}, we propose a definition of an intersection, utilizing the concept of neighboring cells (Definition \ref{def:neighboring}), walls (Definition \ref{def:walls}) and openings (defined as part of Definition \ref{def:intersection}).

\begin{definition}
    \label{def:neighboring}
    The map cell $m_{i_1j_1}$ is neighboring $m_{i_2j_2}$ if $[i_2, j_2]\in \{[i_1+1, j_1],[i_1-1, j_1], [i_1, j_1+1], [i_1, j_1-1]\}$.
\end{definition}

\begin{definition}
    \label{def:walls}
    A set of neighboring unoccupied cells is defined as a wall $W_v$, if and only if, for each map cell $m_{i_1j_1}\in W_v$ there exists an unknown or occupied map cell $m_{i_2j_2}$ where $[i_2, j_2]\in \{\{i_1+1,i_1,i_1-1\}\times \{j_1+1,j_1,j_1-1\}\}$ and the last and first map cell in the set are neighboring each other.
\end{definition}
\begin{definition}
\label{def:intersection}
%$I_i$ is defined as a set of unoccupied cells, where each cell has at least one neighboring cell that belongs to the same set $I_i$ on the grid map $M$. 
Let $I_i$ be a set of unoccupied cells, where each cell has at least one neighboring cell that belongs to the same set $I_i$ on the grid map $M$. $I_i$ is considered an intersection if and only if there exists no wall $W_v$ such that all cells in the set $W_v$ are part of $I_i$ and all cells in the set $I_i$ fulfill the following requirements; each cell in the set can only neighbor i) cells belonging to $I_i$, ii) occupied cells and iii) cells belonging to other sets of connected cells $O_i$, $i\in \{1, 2,\ldots, n\}$ with $n \geq 3$, which represent openings of the intersection. Furthermore, map cells in an opening $O_i$ are not allowed to neighbor a cell belonging to another opening $O_j$.   
\end{definition}
This article presents a methodology to detect intersections (defined in Definition \ref{def:intersection}) in a grid map, which will be used subsequently as branch points in the topometric map. The set of identified intersections $\mathcal{I}$ will be utilized to find paths between intersections, dead ends, and pathways leading to unexplored areas of the map. In this work, our objective is to identify intersections 
%without relying on preconceived notions regarding
with no a priori assumptions on the input map's shape, size, or number of connecting roads or corridors. This objective necessitates a methodology that does not depend on specific features in the environment, unlike many existing methods that look to extract specific features (such as rooms) present only in indoor environments. 

The illustration in Fig.~\ref{fig:algower} presents the overall concept of the proposed algorithmic framework and the main steps involved in detecting intersections and creating a topological map. Details about the individual steps %steps will be further analyzed and 
are presented in sequential subsections. As depicted, the first step in detecting intersections %is performed by
involves analyzing how a given map is connected using the method described in the subsection~\ref{sec:gap}. 
%As such, 
The output from the first subsection is used in subsection~\ref{sec:gapDetections} to find gap detections (potential openings for intersections) and to filter out openings that are too small for the robot to pass. In subsection~\ref{sec:walls}, the output from~\ref{sec:gapDetections} is used to find walls and filter out small objects, following which, in subsection \ref{sec:opDetect}, possible openings into intersections are found. In subsection \ref{sec:opOpt}, the opening detections are moved to resolve overlaps. Then, in subsection \ref{sec:detectInter}, the opening detections are connected to create intersections and pathways. The intersections are optimized in subsections \ref{sec:interOpt}. Finally, using the constructed semantic map, a topological map in the form of a skeleton is generated in subsection \ref{sec:roboPath}. 
\begin{remark}
Although we have defined the notions of walls and intersections, it is difficult to design a systematic and methodical approach to finding such semantic regions in a general map, especially in unstructured and outdoor environments, which are also under consideration in this article. It is for this reason that some of the methods used in the GRID-FAST approach may appear heuristic in nature. We demonstrate the effectiveness of the proposed method through validation on real maps gathered from robotics experiments in structured and unstructured indoor and outdoor environments. 
\end{remark}
\section{Methodology}
\label{sec:Implementation}
%\section{Methodology for Semantic \& Topometric Mapping}
%\section{Implementation}
\subsection{Gap Segmentation} \label{sec:gap}
%%%%%%%%%%%%%%%%%%%%%%%%

In this subsection, taking the occupancy grid $M$ as an input, we derive the 
gaps present in the map denoted as $g_{ik}$, where the index $i \in \{1,\ldots, I\}$ denotes the line of the occupancy grid that the gap belongs to and the index $k \in \{0, 1, \ldots, K\}$ is used to assign a number to the gaps in a given row $i$, with $K$ being the total number of gaps in row $i$.
The gap $g_{ik}$ is defined as a series of connected unoccupied cells on the same row, as represented in Eq. \eqref{eq:gapdef}. 
\begin{equation}
\label{eq:gapdef}
    g_{ik}=\{m_{ig_{ikS}}, m_{i(g_{ikS}+1)},m_{i(g_{ikS}+2)}, \hdots m_{ig_{ikE}}\}
\end{equation}
where the column index of the first and last positions of the cells belonging to the $k^{\text{th}}$ gap of row $i$ are denoted by $g_{ikS}$ and $g_{ikE}$, respectively. It is not uncommon for occupancy maps to have unoccupied cells incorrectly classified as unknown in open spaces, thus creating 'holes' in the map. To allow for some holes in the map, gaps are allowed to contain small groups of maximum $f_{uk}$ number of connected unknown cells, as long as it is not the first or last cell in the gap.

In the sequel, we make the practical assumption that the robot has a finite size $R_{min}$, a positive real number. Then, there is a minimum size of the gap that the robot can pass, captured by the integer $g_{min}=\lceil R_{min}/M_{cell}\rceil$, where $M_{cell}\in\mathbb{R}$ is the cell size of the occupancy grid and $\lceil \cdot \rceil$ denotes the ceiling operation. The gaps can belong to one of two groups: gaps that the robot can traverse $\mathcal{G}_t$ and non-traversable gaps $\mathcal{G}_{nt}$, as defined in Eq. \eqref{eq:gt}. 
\begin{equation}
    \label{eq:gt}
    \begin{matrix}
        \mathcal{G}_t \: \: = & \{g_{ik}\: |\: g_{ikE}-g_{ikS} \geq g_{min}\}\\
        \mathcal{G}_{nt}= & \{g_{ik}\: |\: g_{ikE}-g_{ikS}<g_{min}\}
    \end{matrix}
\end{equation}
It should be noted that gaps that belong to $\mathcal{G}_t$ on the same row cannot overlap or be connected, i.e., $g_{ikS}>g_{i(k-1)E}+1$. Each gap $g_{ik}\in \mathcal{G}_t$ is connected to a group of gaps \blue{$G^-_{ik}$} in the previous line and a group \blue{$G^+_{ik}$} in the next line, which is defined as in Eq. \eqref{eq:groupGapDef}.
\blue{\begin{equation}
\label{eq:groupGapDef}
    \begin{aligned}
        G^+_{ik_1}=\{g_{(I + 1)k}\:|\: &g_{(i + 1)k} \in \mathcal{G}_t,
        g_{(i + 1)kS}<g_{ik_1E},\: g_{(i + 1)kE}>g_{ik_1S}, \\ 
        &\min(g_{(i + 1)kE},g_{ik_1E})-\max(g_{(i \pm 1)kS},g_{ik_1S})\geq g_{min}\}\\
        \text{and}\\
        G^-_{ik_1}=\{g_{(I - 1)k}\:|\: &g_{(i - 1)k} \in \mathcal{G}_t,
        g_{(i - 1)kS}<g_{ik_1E},\: g_{(i - 1)kE}>g_{ik_1S}, \\ 
        &\min(g_{(i - 1)kE},g_{ik_1E})-\max(g_{(i - 1)kS},g_{ik_1S})\geq g_{min}\}
    \end{aligned}
\end{equation}}
If two gaps $g_1$ and $g_2$ on the line $i$ and $i\pm 1$ overlap, and the overlapping sections, denoted as $g_{o1}$ and $g_{o2}$ respectively, have a size less than the specified threshold $g_{min}$, then the overlapping region can not be traversed by a robot of size $R_{min}$ and the overlap is disregarded. Instead, the cells from $g_{o1}$ and $g_{o2}$ are 
treated as two new gaps, removed from $g_1$ and $g_2$ respectively and made part of the $\mathcal{G}_{nt}$ collection.

\subsection{Gap Detections and Opening Filtering} \label{sec:gapDetections}
\begin{figure}[htbp]
    \centering
    \includegraphics{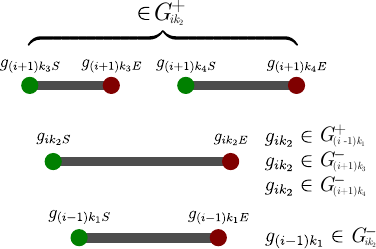}
    \caption{An example of how the gaps $g_{(i-1)k_1}$, $g_{ik_2}$, $g_{(i+1)k_3}$, and $g_{(i+1)k_4}$ could be connected. Here, \blue{$G^+_{ik_2}$} has two elements and is treated as a gap detection.}
    \label{fig:gapOver}
\end{figure}

A gap detection is a \blue{$G^+_{ik}$ or $G^-_{ik}$} collection that contains two or more gaps\blue{, i.e. the set of gap detections is defined as:
\begin{equation}
    \mathcal{D}=\{G^+_{ik} / G^-_{ik} \;|\; \text{card}\{G^+_{ik} / G^-_{ik}\}\geq 2\}
\end{equation}}
A gap detection indicates a potential opening into an intersection, as gaps are points in the map where two or more paths intersect. Fig. \ref{fig:gapOver} illustrates an example of this concept.

\begin{figure}[htbp]
    \centering
    \begin{subfigure}[b]{0.243\linewidth}
         \centering
         \includegraphics[width=\textwidth]{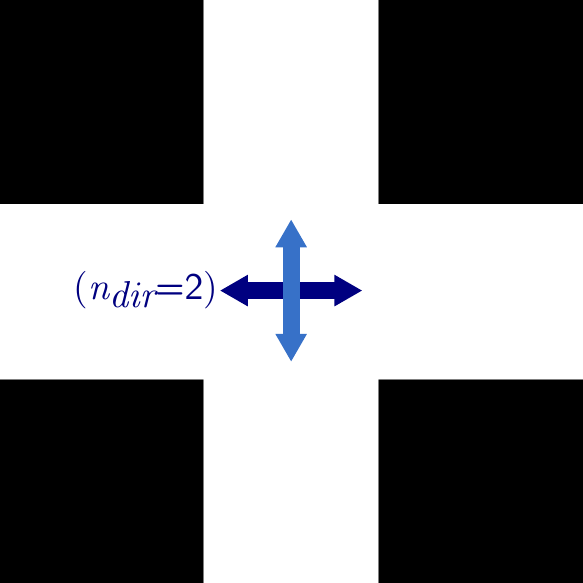}
         \caption{$n_{dir}=1$, $n_{dir}=2$}
         \label{fig:scanAng1&2}
     \end{subfigure}
     \hfill
     \begin{subfigure}[b]{0.243\linewidth}
         \centering
         \includegraphics[width=\textwidth]{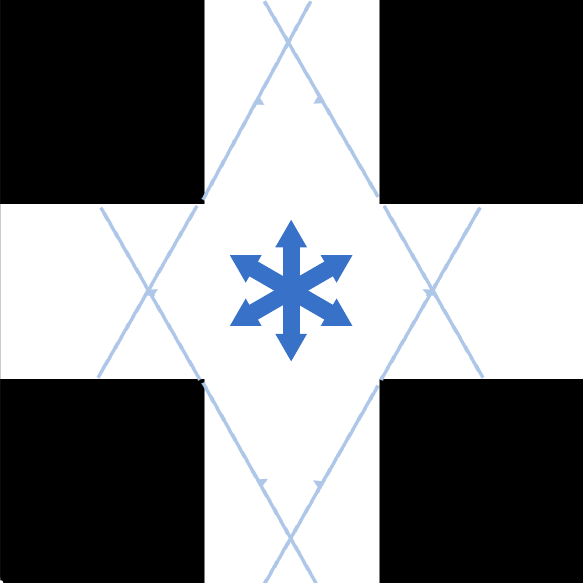}
         \caption{$n_{dir}=3$}
         \label{fig:scanAng3}
     \end{subfigure}
     \begin{subfigure}[b]{0.243\linewidth}
         \centering
         \includegraphics[width=\textwidth]{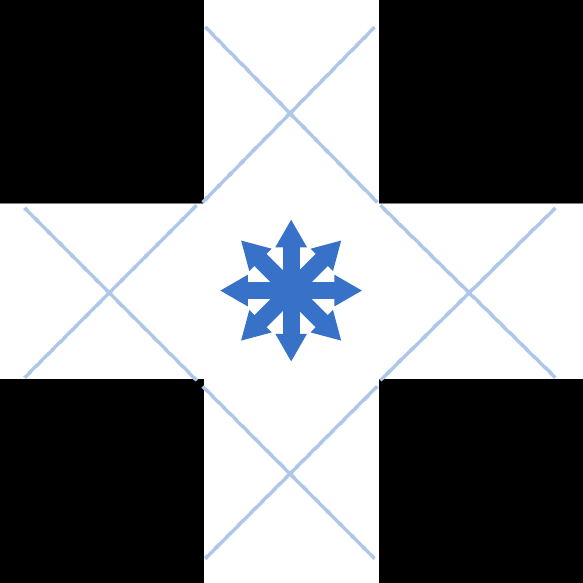}
         \caption{$n_{dir}=4$}
         \label{fig:scanAng4}
     \end{subfigure}
     \begin{subfigure}[b]{0.243\linewidth}
         \centering
         \includegraphics[width=\textwidth]{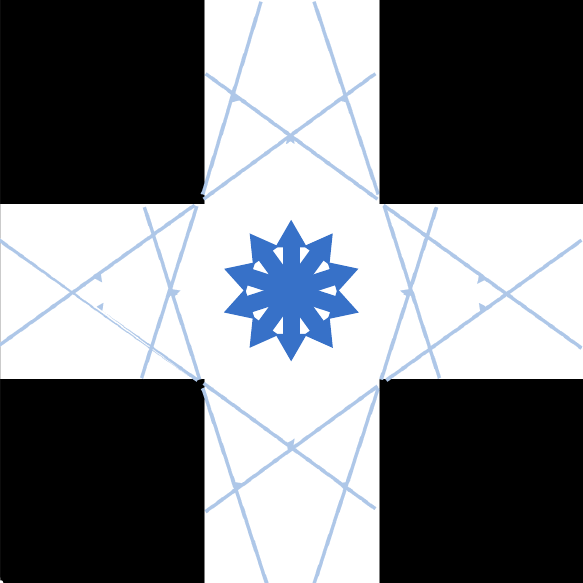}
         \caption{$n_{dir}=5$}
         \label{fig:scanAng5}
     \end{subfigure}
    \caption{A perfect cross intersection aligned with the first scanning angle. The figures show the result of using different numbers of scan directions. The gap detection groups are indicated by blue lines and the arrows indicate the directions the image was scanned in.}
    \label{fig:scanAng}
\end{figure}
A drawback of performing the gap segmentation just once is that it cannot detect all openings in all scenarios. A good example of one such situation is a perfect cross intersection that is perpendicular to the scanning direction, where no gap detections are found, as the number of connected gaps for each row is always one, as shown in Fig. \ref{fig:scanAng1&2}. To solve this problem, the occupancy map is scanned $n_{dir}$ times using the gap segmentation in subsection \ref{sec:gap}, where $n_{dir}$ is a positive integer. Each time, the input map is rotated by $\theta=\pi /n_{dir}$ radian and segmented into gaps. Fig. \ref{fig:scanAng} shows the resulting 
gap detections for different values of $n_{dir}$.

\begin{figure*}
    \centering
    \begin{subfigure}[b]{0.24\textwidth}
         \centering
         \includegraphics[width=\textwidth]{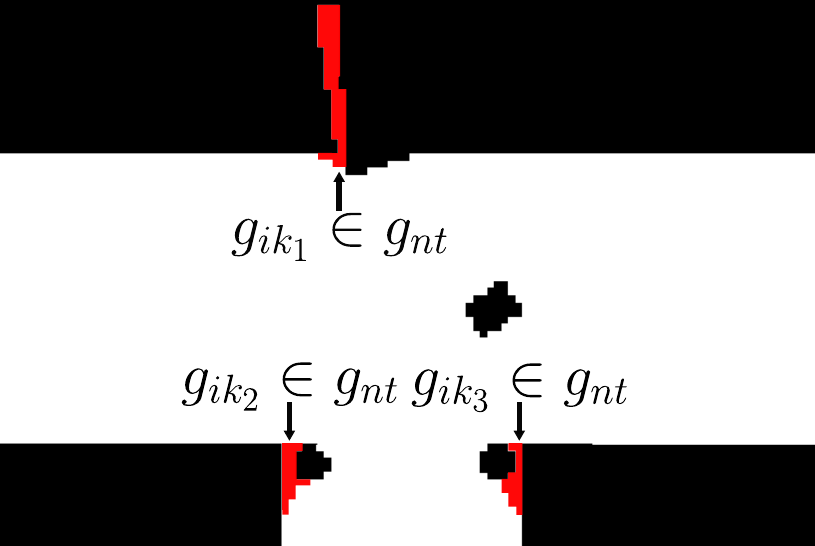}
         \caption{}
         \label{fig:filterdA}
     \end{subfigure}
     \hfill
     \begin{subfigure}[b]{0.24\textwidth}
         \centering
         \includegraphics[width=\textwidth]{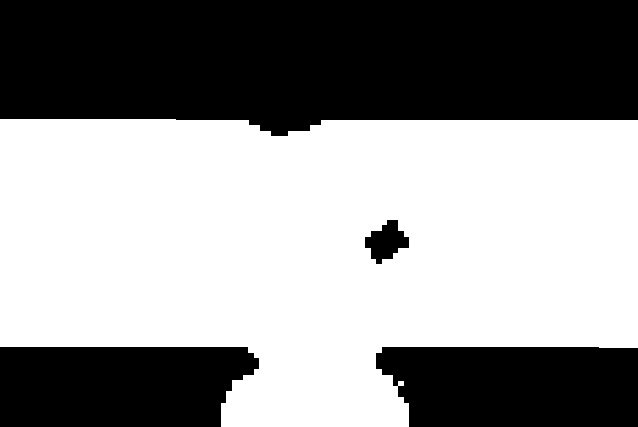}
         \caption{}
         \label{fig:filterdB}
     \end{subfigure}
     \begin{subfigure}[b]{0.24\textwidth}
         \centering
         \includegraphics[width=\textwidth]{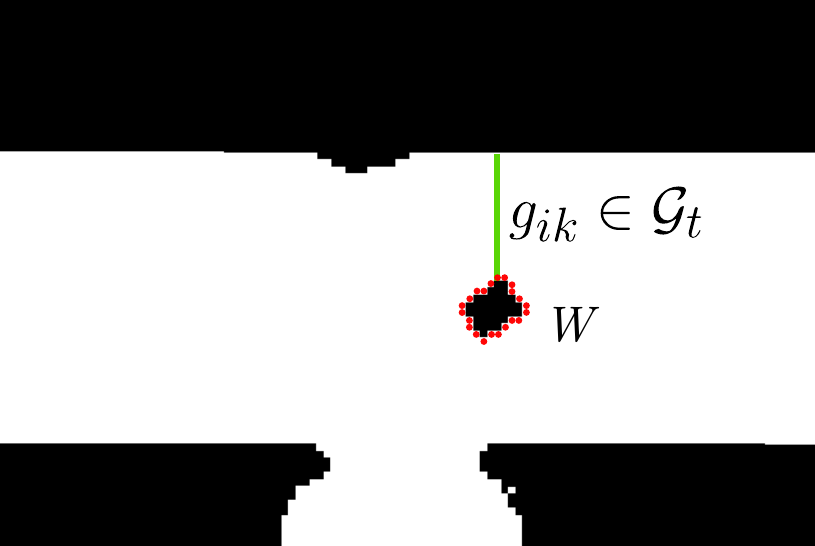}
         \caption{}
         \label{fig:filterdC}
     \end{subfigure}
     \begin{subfigure}[b]{0.24\textwidth}
         \centering
         \includegraphics[width=\textwidth]{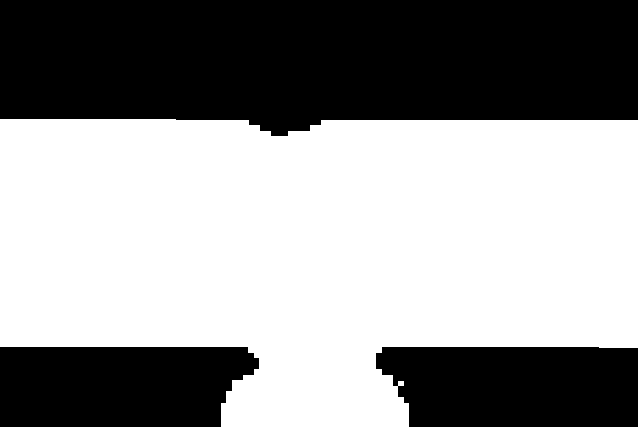}
         \caption{}
         \label{fig:filterdD}
     \end{subfigure}
    \caption{(a) Unfiltered occupancy map where some of the gaps belonging to $\mathcal{G}_{nt}$ are marked. (b) Filtered occupancy map where small openings have been removed. (c) Map showing a gap part of the set $\mathcal{G}_t$ that is connected to an object. The points on the wall $W_n$ around the object have fewer cells with a neighboring occupied point than the threshold $f_{obj}$, and therefore the object is removed. (d) Final filtered map where small objects have been removed.}
    \label{fig:filter}
\end{figure*}

The gaps in the group $\mathcal{G}_{nt}$, defined in Eq. \eqref{eq:gt}, are used to remove areas in the map that the robot cannot traverse. This is achieved by setting all $m_{ij} \in g_{ik} \in \mathcal{G}_{nt}$ to occupied if $g_{ik}$ is next to occupied cells. Otherwise, the unoccupied cells contained in $g_{ik}$ are set to unknown, as shown in Fig. \ref{fig:filterdA} and \ref{fig:filterdB}. 
\subsection{Wall Detection and Object Filtering} \label{sec:walls}
As given by Definition \ref{def:intersection}, the intersection area is defined by its openings and the walls between them. Because of this, it is crucial to include the identification of walls as part of the solution. This method represents a wall as a set of map cells defined in Definition \ref{def:walls}. 

Each cell in the wall corresponds to a cell on the edge of the unoccupied spaces of the map, and there is a distance of one cell length between each cell in the collection, including the first and last cell.
The method to detect walls utilizes the fact that all spaces that either have a concave shape or contain at least one object inside of it will have a gap detection connected to each wall (assuming that the number of scanning directions, \( n_{dir} \), in subsection \ref{sec:gapDetections} is high enough). This would mean the proposed wall detection method would not work for simple cases such as a plain square or circular room with no objects. This is not of concern in the context of this article, as creating a topometic map in such convex spaces is trivial. Therefore, we do not consider such scenarios in this article.

The wall cells are thus gathered by following the edges of the unoccupied space at the start and end point of each gap, which is part of the gap detection until all the wall cells on the map have been detected. The detected walls are used to remove small objects and unoccupied cells classified as occupied cells from the map. An object is removed if the wall around the object has fewer cells with a neighboring occupied point than the threshold $f_{obj}$, which is a positive integer. The object is removed using a polygon filling algorithm \cite{Dunlavey1983}, as shown in Fig. \ref{fig:filterdC} and \ref{fig:filterdD}. All gaps connected to the wall are removed and are not considered in the later steps of the method. 

Small objects are removed as they introduce unnecessary complexity to the map. However, this poses a problem, as the global paths generated in the subsequent subsection \ref{sec:SkeletonGen} might overlap with objects, necessitating that the robot possesses a local navigation solution to navigate around small objects in its path. If this is not the case, the value $f_{\text{obj}}$ can be set to a small value, which leads to the removal of only noise from the map at the expense of increased complexity in the topometric map. 
\subsection{Opening Detection} \label{sec:opDetect}
\begin{figure}[h]
    \centering
    \includegraphics[width=.6\linewidth]{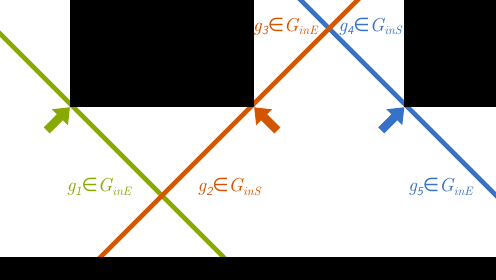}
    \caption{A part of a map with three gap detections is shown. The direction of the gap detection and the inside points of the gaps are marked by the arrows in the figure. In this situation, there is an opening between the inside point of $g_3$ and $g_4$. There will not be an opening between the inside points of $g_1$ and $g_2$, as they do not fulfill the second condition P2 in sub-section \ref{sec:opDetect} and thus will be converted directly to opening detections. $g_5$ will be converted to an opening detection as it does not overlap any other gap detections.}
    \label{fig:opDetect}
\end{figure}
An opening detection $o_t$ is an opening in and out of an intersection. Opening detection is defined as a pair of points $o_t = \{ m_{i_1j_1}\in W_{v_{1}}, m_{i_2j_2}\in W_{v_{2}}\}$, which can be part of one or two walls $W_{v_{1}}$ and $W_{v_{2}}$.
The gap detections from Subsection \ref{sec:gapDetections} are used to identify the opening detections. Using the inside point $g_{ikI}$ of gaps, as defined in Definition \ref{def:insidePoints}, potential openings to intersections are found in the map. 

\begin{definition}
    \label{def:insidePoints}
    All gaps $g_{ik}$\blue{, that are part of a gap detection, \blue{i.e. $g_{ik}\in G^+_{ik} / G^-_{ik} \in \mathcal{D}$,} as defined in Subsection \ref{sec:gapDetections}, have at least one inside point $g_{ikI}$. The inside point} can either be the gap's start point $g_{ikI}=\{g_{ikS}\}$, end point $g_{ikI}=\{g_{ikE}\}$, or both, $g_{ikI}\in\{g_{ikS},g_{ikE}\}$. For the case where a gap collection contains a series of gaps \blue{$G^+_{ik_1}=\{g_{(i+1)k_2},g_{(i+1)(k_2+1)},\dots,g_{(i+1)(k_2+N)}\}$ or $G^-_{ik_1}=\{g_{(i-1)k_2},g_{(i-1)(k_2+1)},\dots,g_{(i-1)(k_2+N)}\}$}, i) if a gap is the first gap in the gap detection, then the inside point is $g_{ikI}=\{g_{ikE}\}$, ii) if it is the last then $g_{i(k+N)I}=\{g_{i(k+N)S}\}$ iii) else $g_{i(k+n)I}=\{g_{i(k+n)S},g_{i(k+n)E}\}$ for $0<n<N$.
\end{definition}
These inside points are crucial as they are situated at points where paths meet in an intersection, as depicted in Figure \ref{fig:opDetect}.
Endpoints not part of $g_{ikI}$ are not always guaranteed to be within the intersection. To enhance the accuracy of intersection detection, gaps are combined to find opening detections between their inside points. A potential issue arises when two gaps, each a part of different intersections, are combined, which may result in an inaccurate opening. To address this, the following method is utilized.
A gap, part of a gap detection, can be in one of three collections, $G_{inS}$, $G_{inE}$ or $G_{inB}$, depending on whether its start or end point is an inside point and if it is part of a \blue{$G^+_{ik}$ or $G^-_{ik}$} gap detection, \blue{where $G^+_{ik}$ and $G^-_{ik}$, defined in Subsection \ref{sec:gap}, are the directions of detection compared to the gap that is connected to the gap detection, i.e., if the gap detection was found on a previous or the next line to the gap.}
Table \ref{tab:gapDetectType} shows different collections to which a gap can belong. 
\begin{table}[h]
\caption{The gaps (part of a gap detection) can be classified into three categories $\{G_{inS}, G_{inE}, G_{inB}\}$ depending on the gap detection direction and the inside point of the gap detection.}
\label{tab:gapDetectType}
    \centering
    \begin{tabular}{rlc|cc|}
        \clineorig{4-5}
        
        &&&\multicolumn{2}{c|}{Detection direction} \\ \clineorig{4-5}
        &&& \multicolumn{1}{c|}{$G^+_{ik}$} & $G^-_{ik}$  \\ \clineorig{1-5}
        \multicolumn{1}{|r}{\multirow{3}{*}{\begin{turn}{90} Inside \end{turn}}}
        &\multirow{3}{*}{\begin{turn}{90} point \end{turn}}
        &\multicolumn{1}{|c|}{$g_{ikI}=\{g_{ikS}\}$ }& \multicolumn{1}{c|}{$G_{inS}$ \cellcolor{LightBlue} } & $G_{inE}$ \cellcolor{LightBlue}\\ \clineorig{3-5}
        \multicolumn{1}{|r}{}&&\multicolumn{1}{|c|}{$g_{ikI}=\{g_{ikE}\}$} & \multicolumn{1}{c|}{$G_{inE}$ \cellcolor{LightBlue}} & $G_{inS}$ \cellcolor{LightBlue}\\ \clineorig{3-5}
        \multicolumn{1}{|r}{}&&\multicolumn{1}{|c|}{$g_{ikI}=\{g_{ikS},g_{ikE}\}$}& \multicolumn{1}{c|}{$G_{inB}$ \cellcolor{LightBlue}} & $G_{inB}$ \cellcolor{LightBlue}\\ \hline
    \end{tabular}
   % \caption{Which collection a gap, part of a gap detection, belongs to}
    
\end{table}

% An alternative table proposal
%\begin{table}[h]
%\caption{The gaps (part of a gap detection) can be classified into three categories $\{G_{inS}, G_{inE}, G_{inB}\}$ depending on the gap detection direction and the inside point of the gap detection.}
%\label{tab:gapDetectType}
%    \centering
%\begin{tabular}{|c|c|c|}
%  \hline
%  Inside Point $(g_{ikI})$ & Detection Direction $(G_{i \pm 1})$ & Gap Category $(G_{inS}, G_{inE}, G_{inB})$ \\
%  \hline
%  $\{g_{ikS}\}$ & $G_{i+1}$ & $G_{inS}$ \\
%  \hline
%  $\{g_{ikS}\}$ & $G_{i-1}$ & $G_{inE}$ \\
%  \hline
%  $\{g_{ikE}\}$ & $G_{i+1}$ & $G_{inE}$ \\
%  \hline
%  $\{g_{ikE}\}$ & $G_{i-1}$ & $G_{inS}$ \\
%  \hline
%  $\{g_{ikS},g_{ikE}\}$ & $G_{i+1}$ & $G_{inB}$ \\
  %\hline
  %$\{g_{ikS},g_{ikE}\}$ & $G_{i-1}$ & $G_{inB}$ \\
%  \hline
%\end{tabular}
%\end{table}

\blue{In Table 1, gaps (part of a gap detection) are classified into one of the following three categories $\{G_{inS},G_{inE},G_{inB}\}$. The classification rule is based on the gap detection direction, which has two possibilities $\{G^+_{ik}, G^-_{ik}\}$ and the inside point which has three possibilities $\{g_{ikS}, g_{ikE}, \{g_{ikS}, g_{ikE}\}\}$. For instance, if a gap detection $G_i$ has $g_{ikS}$ as the inside point and $G^-_{ik}$ as the gap detection direction, then it is classified into the category $G_{inE}$.} 

Frontiers need to be identified to generate the topometric maps for partially explored maps. A frontier, denoted as $f_c$, is a set of connected cells with at least one unknown neighboring cell. The distance between the first and last cell in $f_c$ equals or exceeds $R_{min}$. The cells in the frontier are also part of a wall, as they satisfy the conditions presented in subsection \ref{sec:walls}. Therefore, $f_c$ is a subset of $W_v$. Once identified, all gaps that have an inside point connected to a wall cell, which is part of the frontier, are removed. Subsequently, an opening detection is placed at the frontier's start and end points.

If two different gaps overlap each other, there is an opening between their respective inside points if:
\begin{enumerate}
    \item[P1.] One of the gaps belongs to  $G_{inS}$, and the other belongs to $G_{inE}$.
    \item[P2.] The intersection point $p$ is behind the line from the inside point of the gap in $G_{inS}$ to the inside point of the gap in $G_{inE}$, i.e., the condition in \eqref{eq:openingCon} is true, where $N_o$ is the normal to the line and $V_o$ is the vector between the inside point of the gap part of $G_{inS}$ to $p$. 
    \begin{equation}
    \label{eq:openingCon}
    N_o \cdot V_o<0
    \end{equation}
    \item[P3.] The line between the gaps' inside points does not overlap a wall. 
    \item[P4.] If a gap is overlapping multiple gaps that satisfy the above conditions, then only that gap that results in the smallest opening possible \blue{for both gaps} is used.
\end{enumerate}

If a gap does not overlap with a gap that satisfies the above conditions or is part of the $G_{inB}$ collection, it is converted to an opening as long as none of the ends of the gap is connected to a frontier. In Figure \ref{fig:opDetect}, we present a case to illustrate the method of opening detection. 
\subsection{Resolution of Overlapping Openings} \label{sec:opOpt}
All the opening detections in subsection \ref{sec:opDetect} are detected without considering other openings, which results in overlapping openings (as can be seen in the opening detection block of Figure \ref{fig:algower}). As per Definition \ref{def:intersection}, openings to intersections can not share the same space. Therefore, it is necessary to ensure that opening detections do not overlap.
When two opening detections $o_{t_1}$ and $o_{t_2}$ overlap, the solution involves swapping the positions of their respective points. This method is applicable when the openings share a common wall, i.e., both are contained within $W_v$. If there are multiple points forming the openings on the same wall or if they share two common walls, the point with the fewest cells between them is chosen for the swap. However, if the openings do not share a wall, the overlap is resolved by removing the longest opening. 
\subsection{Intersection Detection and Pathway Detection}
\label{sec:detectInter}
\begin{figure}[t]
    \centering
    \includegraphics[width=.4\linewidth]{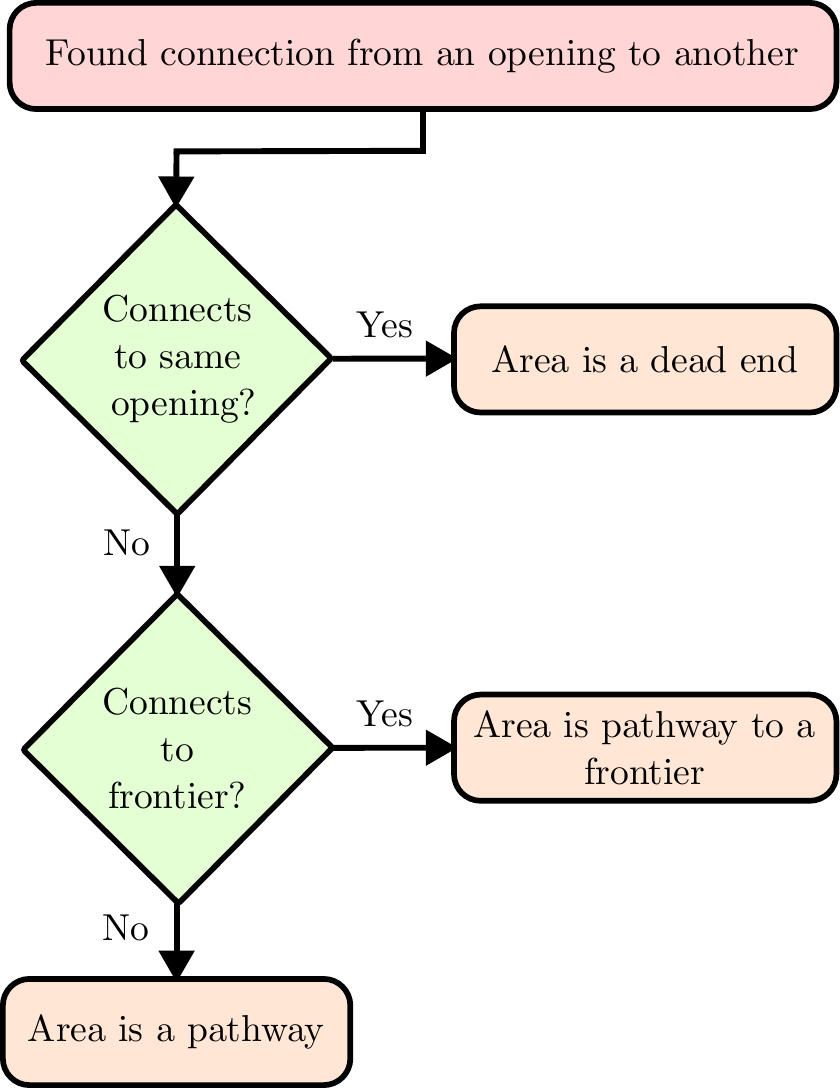}
    \caption{Method to classify pathways between intersections into regular pathways, pathways leading to frontiers and dead ends. }
    \label{fig:pathchart}
\end{figure}

Intersections are areas enclosed by opening detections and the walls connecting them. Each opening detection has two potential intersection openings, one facing each direction. The intersection identification is made by selecting an opening that is not part of an already detected intersection and following the wall from one of the ends of the opening until the next opening. The process is then repeated with a new opening until a connection between the last opening and the first is found, thus closing the loop. The area of an intersection is the area enclosed by the intersection's openings and the points along the walls connecting these opening detections. If an intersection contains less than three openings, the intersection and its opening are removed. This approach finds intersections consistent with Definition \ref{def:intersection}, which presents conditions for a group of cells to be called an intersection.   
A method similar to the one used to detect intersections is used to find the pathways between them. However, instead of following the walls into the intersection, they are followed away from it. As a result, if two openings are linked, a pathway is established between them. If an opening is linked with a frontier, the area is classified as a pathway leading to an unexplored region. During the wall search, if the wall being followed leads back to the original opening, the area is classified as a dead end. This process of classifying pathways is illustrated in Figure \ref{fig:pathchart}.
\subsection{Intersection Optimization} \label{sec:interOpt}
The intersections identified in the previous section may not be optimal, as they can be unnecessarily large or too small. The openings in the intersections might not be placed in optimal positions. For instance, they could be placed at odd angles or in the widest part of the opening, as seen in the intersection detection step in Fig. \ref{fig:algower}. Additionally, some intersections may be unnecessary as some openings are not connected to actual paths. For example, the method can sometimes interpret a corner of a large room as a pathway. These intersections add unnecessary complexity to the topometric map.

As the intersection is constructed through its openings, it is crucial to note that each opening of the intersection, denoted as $o_{In}$, where $n$ is the index of the opening of intersection $I$, connects two walls $W_{Sn}$ and $W_{En}$ at $o_{Ins}$ and $o_{Ine}$ respectively. It is also important to remember that every opening in the intersection shares each wall with another opening in the same intersection. Hence, $W_{Sn}=W_{E \mod (n+1,I_o)}$, where $I_o$ represents the number of openings in the intersection. 
Let us denote by $\mathcal{W}_I$ all possible configurations of $I$, which are all possible combinations of its openings and their potential connections to their respective walls (combinations where openings do not overlap other openings or a wall).
Next, to make a choice among the possibilities, we introduce the following optimization problem:
\begin{equation}
\label{eq:InterMin}
\begin{matrix}
    \min_{\mathcal{W}_I} \left( \sum^{I_o}_{n=0}\left( || o_{In}|| +D(d_c) \right) \right) \\
     \\
 \text{where}\;\;   D(x)=\left\{ \begin{matrix}
        D_{max}-\frac{D_{max}}{d_{cMin}}*x & x\leq d_{cMin} \\
        (x-d_{cMin}) & x > d_{cMin}
    \end{matrix} \right.
\end{matrix}.
\end{equation}
For the GRID-FAST method, an optimal intersection is the one that minimizes the cost function in the optimization problem \eqref{eq:InterMin}, where the value $d_c$ represents the distance from the center of the opening to the center of the intersection. In the definition of $D(x)$, $d_{cMin}$ is the value of $d_c$ where $D(x)$ attains it's minimum, and $D_{max}$ is the maximum penalty for choosing too small a value for $d_c$ (attained at $d_c=0$). 
The goal of the minimization problem is to position the intersections' openings in the smallest part of the pathway leading to the intersections, aligned with the direction of the pathway. This is achieved by placing the opening such that it is as small as possible, hence the term $||o_{In}||$ in the optimization problem \eqref{eq:InterMin}. However, it is detrimental if the openings are positioned too far from the intersection or too close, as this will result in unnecessarily large or small intersections. This concern is addressed by the additive term \( D(d_c)\) in the optimization problem \eqref{eq:InterMin}. It should be noted that both $||o_{In}||$ and $d_c$ depend on the position of $o_{In}$, the set of openings to an intersection.

Each intersection is optimized in a two-step search. First, the smallest possible intersection is found, and then it is grown to minimize the cost function in \eqref{eq:InterMin}. During the optimization of a particular intersection, all other intersections are static
to ensure that the openings of the optimized intersection cannot be placed within another intersection or after a frontier. Consequently, the part of the wall at which the openings in $I$ can be placed corresponds to cells belonging to $w_{ns}\subset W_{ns}$, where the wall-section $w_{ns}$ (part of the complete wall $W_{ns}$) originates from a neighboring intersection or frontier  passes through $I$, and continues to the subsequent intersection or frontier. A particular scenario arises when the intersection being optimized is connected to a dead end or the path loops back to another opening of the same intersection, where one of the two neighboring intersections is the currently optimized intersection. In this case, 
the first or the last half of the points from the neighboring intersection to the current intersection are removed from $w_{ns}$. This step addresses issues that might arise when 
a path loops back to the original intersection and eliminates the solution where one of the openings faces a wall (in the case of a dead end).

The openings are optimized pairwise to find the smallest intersection. The start point of $o_{In}$ and the end point of the next opening ($o_{I\mod(n+1,I_o)}$) are fixed while the wall cells in each $w_{ns}$ are searched for the point $w_{k}\in w_{ns}$ that is the solution to the following optimization problem:
\begin{equation}
    \min_{w_{ns}}(||o_{Ins}\rightarrow w_k||^2 + ||o_{I\mod (n+1,I_o)e}\rightarrow w_k||^2)
    \label{eq:shringInter}
\end{equation}
where the notation $||o_{Ins}\rightarrow w_k||$ denotes the distance between $o_{Ins}$ and $w_k$. Once the point $w_k$ that minimizes the optimization problem \eqref{eq:shringInter} has been located, the positions of the openings $o_{In}$ and $o_{I \mod(n+1,I_o)s}$ are moved to that position. The process is repeated for all openings of $I$. In the next stage, the function \eqref{eq:InterMin} is minimized using the center point of the shrunken intersection. Starting from this position, the openings are moved outward, growing the intersection to find the optimal intersection. By first shrinking and then growing the intersection, the resulting intersection will not have any overlapping openings.
\begin{algorithm}[b]
\caption{A method to grow a given intersection. The method takes as input an intersection that is minimized through the optimization in \eqref{eq:shringInter} and outputs an intersection that minimizes the objective in the problem \eqref{eq:InterMin}. }\label{alg:growInt}
\begin{algorithmic}
\scriptsize
\For{each $n\in[0,I_0-1]$}
    \State $ws[i] \gets w_{ns(i_0-i)} \text{ for } i \in [0, I-1]$
    \State $we[j] \gets w_{\mod(n+1,I_0)s(j_0+j)} \text{ for } j \in [0, J-1]$
    \State $D_{min} \gets ||w_{\mod(n+1,I_0)sj_0}\rightarrow w_{nsi_0}||$
    \While{$True$}
        \State $D_{1} \gets ||ws[i+1]\rightarrow we[j]||$ \Comment{Distance $D_{1}$}
        \State $D_{2} \gets ||ws[i]\rightarrow we[j+1]||$
        \Comment{Distance $D_{2}$}
        \If{$i+1<I$ and $(j+1\geq J \text{ or } D_{1}<D_{2})$}
            \State $i \gets i+1$
        \ElsIf{$j+1<J$}
            \State $j \gets j+1$
        \Else
            \State break
        \EndIf
        \State $D \gets ||ws[i]\rightarrow we[j]||$
        \Comment{Distance $D$}
        \If{$D<D_{min}$}
            \State $D_{min} \gets D$
            \Comment{The minimum distance $D_{min}$}
            \State $D_{sCount} \gets 0$
            \Comment{Count of steps since last min}
        \EndIf
        \If{$o_{In}\notin O_{end}$ or $D_{sCount}\geq D_s$}
            \State $o_{In} \gets o_{best}$
            \State $D_{sCount} \gets 0$
        \Else
            \State  $D_{sCount} \gets D_{sCount}+1$
        \EndIf

        \State $score \gets D +D(ws[i], we[j])$ \Comment{Using Eq. \eqref{eq:InterMin}} 
        \If{$score<bestScore$}
            \If{$\{ws[i], we[j]\} \in I\in \mathcal{W}$}
            \State $bestScore \gets score$
            \State $o_{best} \gets \{ws[i], we[j]\}$
            \EndIf
        \EndIf
    \EndWhile

    \If{no new $o_{In}$ is found}
        \State Remove $o_{In}$
        \If{$I_0<3$}
            \State Remove $In$
        \EndIf
    \EndIf
\EndFor
\end{algorithmic}
\end{algorithm}

Algorithm \ref{alg:growInt} is used to grow the intersection. Each opening is moved along the wall to find openings that minimize \eqref{eq:InterMin}. To reduce the number of configurations tested, the opening is moved along one of its ends first and then along the other, with the side selected based on the resulting smallest opening. This process is repeated until there are no more points to move. A peculiar scenario arises in the process of growing an intersection when an opening leads to a dead end, denoted as $o_{In} \in O_{\text{end}}$. 
As the search for the optimal opening position proceeds into a dead-end, the lengths of the openings shrink to zero (as the ends of the openings get progressively closer). Therefore, all opening configurations from the end of the dead-end to an opening configuration with the last known minimum length are ignored. To handle noisy environments, there must be at least $D_s$ steps between the minima, which must be counted as a new minimum. If a new position is not found for an opening, it is removed. If the number of openings in the intersection is less than three, the intersection is removed. If the number of openings in the intersection is more or equal to three, the intersection is re-optimized using this section's method. The steps of this method are captured in Algorithm \ref{alg:growInt}.
\subsection{Skeleton Generation for Topological mapping}
\label{sec:SkeletonGen}
\label{sec:roboPath}
The skeleton representation is generated as a collection of robot paths. The robot paths are generated independently in every path area and intersection area. In intersections, the robot paths are generated from the center of each opening to the centroid of the intersection, which is the average of all points of the intersection. In paths between intersections, the robot path is generated between the two openings of the pathway. 

Initially, the route between the start and end point is evaluated for any obstructions. This is accomplished by utilizing three rays: one between the start and end points and the other two parallel to the first ray but offset by $-R_{min}/2$ and $R_{min}/2$, respectively. The path is deemed obstructed if any of the rays overlap with a wall. If the path is unobstructed, the robot path is generated as a straight line between the points. In cases where the path between the start and the end position is obstructed, a local Voronoi graph is generated inside the path area or intersection area between the points, using a thinning algorithm by \cite{Zhang1984}. The method described in subsection \ref{sec:walls} that is used to remove small objects is used here to fill in the area, and then the thinning algorithm is utilized on the filled area to approximate the Voronoi graph. The start and end points are locked during the thinning process, resulting in a direct path between the two points. All other Voronoi paths generated are discarded. In the event of an intersection, the Voronoi pathway is traced from the opening until the direct pathway to the center is unobstructed. Then, the rest of the Voronoi path is replaced with a straight line to the center of the intersection. In instances of dead-ends and pathways that lead to unexplored areas, a Voronoi diagram is generated for that region. All Voronoi pathways, except for the longest one connected to the dead-end opening, are discarded. At this point, a complete topological map of the environment has been created and represented as a skeleton map.
%
%\FloatBarrier
\section{Validation and comparison with the state-of-the-art} \label{sec:comparison}
\begin{figure}[htbp]
    \centering
    \includegraphics[width=0.4\linewidth]{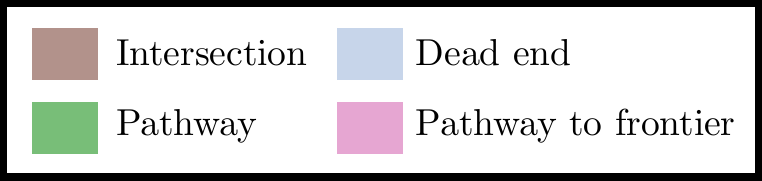}
    \caption{Color scheme used in depicting different semantic regions in the topometric maps generated by GRID-FAST.}
    \label{fig:legend}
\end{figure}

In this section, we present the validation of the GRID-FAST semantic topometric mapping solution. In all the results presented in this section, the topometric map generated by GRID-FAST is overlayed on the filtered occupancy map produced by GRID-FAST using methods in sub-sections \ref{sec:gapDetections} and \ref{sec:walls}. Different semantic regions are represented using the color scheme depicted in Figure \ref{fig:legend}. Additionally, the topological map generated by the method presented in Subsection \ref{sec:SkeletonGen} is displayed as red lines.
\subsection{Parameter Tuning} \label{sec:tuning}
\begin{figure*}[htbp]
    \centering
    \begin{subfigure}[b]{0.24\textwidth}
         \centering
         \includegraphics[,width=\textwidth]{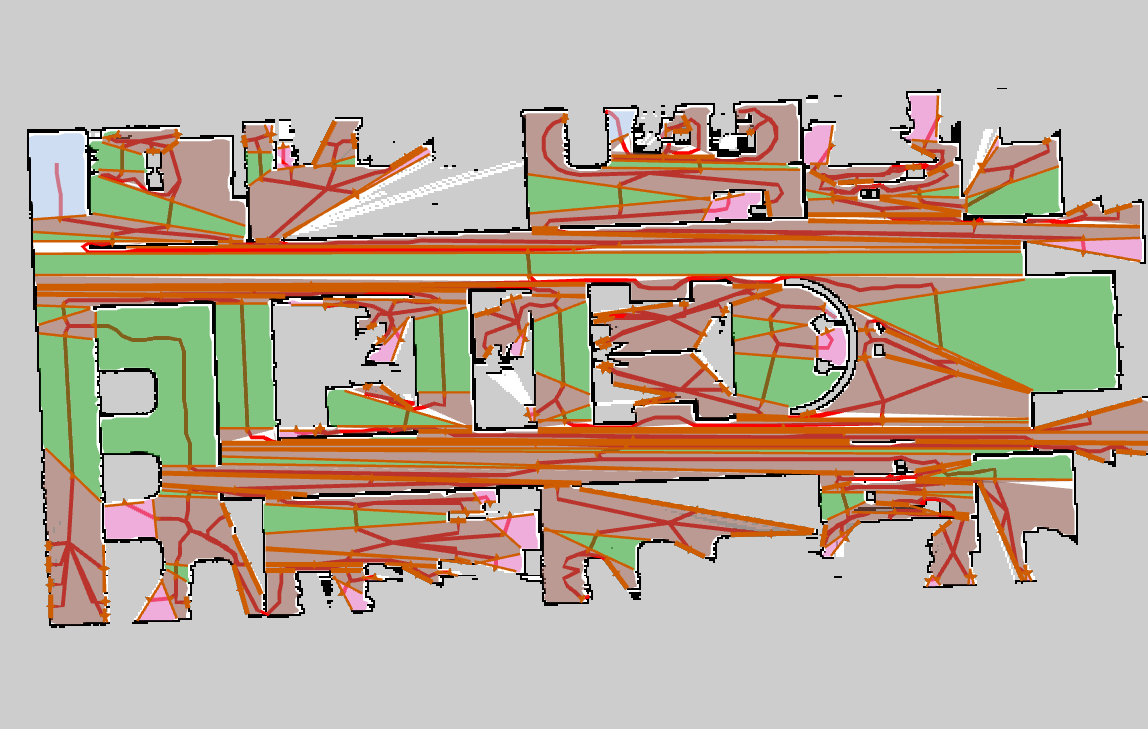}
         \caption{$n_{dir}=1$}
         \label{fig:opt1}
     \end{subfigure}
     \begin{subfigure}[b]{0.24\textwidth}
         \centering
         \includegraphics[clip,width=\textwidth]{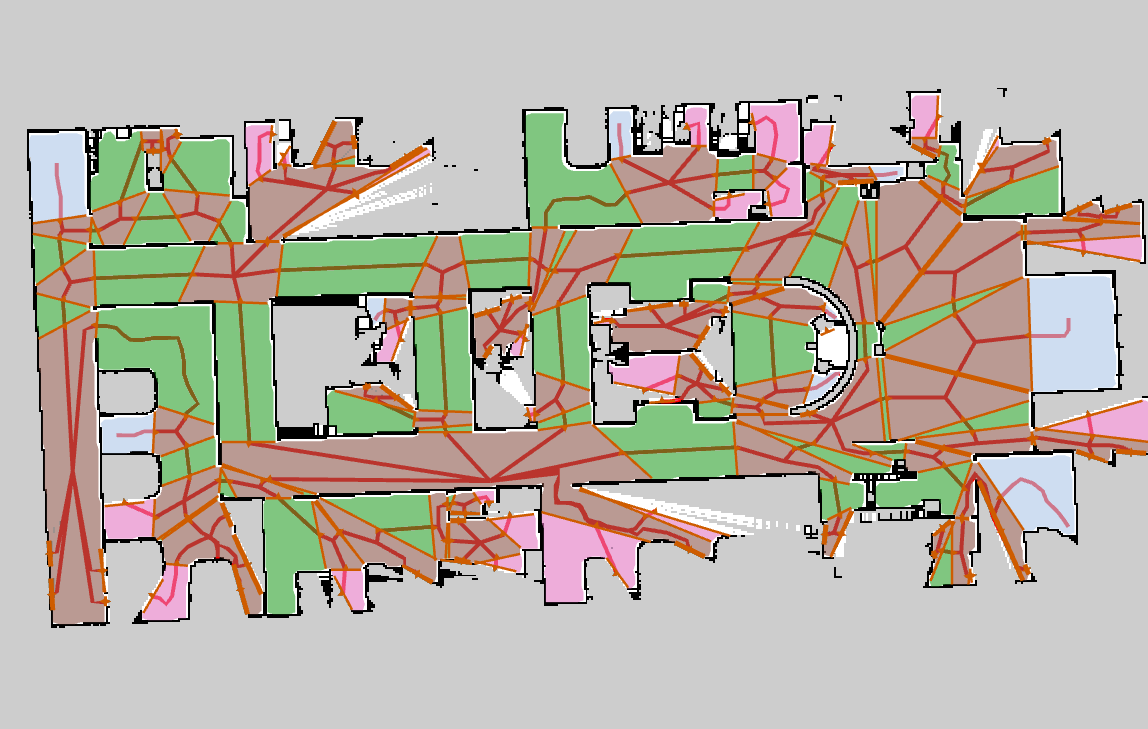}
         \caption{$n_{dir}=2$}
         \label{fig:opt2}
     \end{subfigure}
     \begin{subfigure}[b]{0.24\textwidth}
         \centering
         \includegraphics[width=\textwidth]{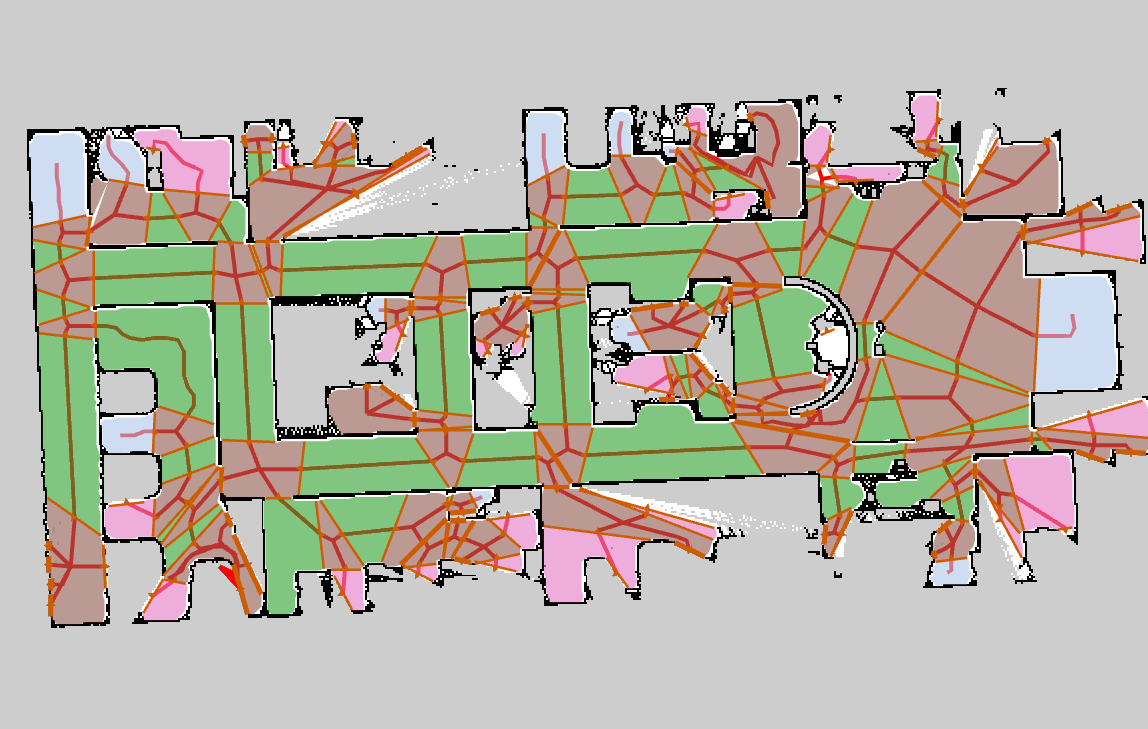}
         \caption{$n_{dir}=3$}
         \label{fig:opt3}
     \end{subfigure}
     \begin{subfigure}[b]{0.24\textwidth}
         \centering
         \includegraphics[width=\textwidth]{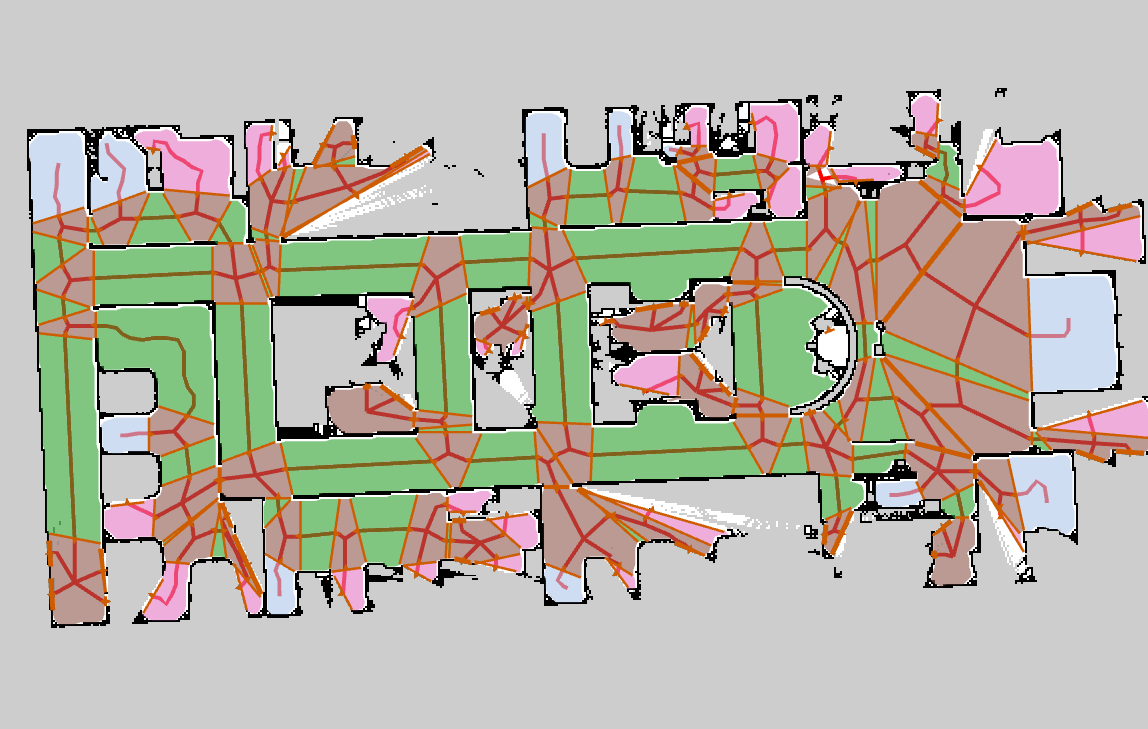}
         \caption{$n_{dir}=4$}
         \label{fig:opt4}
     \end{subfigure}
     \begin{subfigure}[b]{0.24\textwidth}
         \centering
         \includegraphics[width=\textwidth]{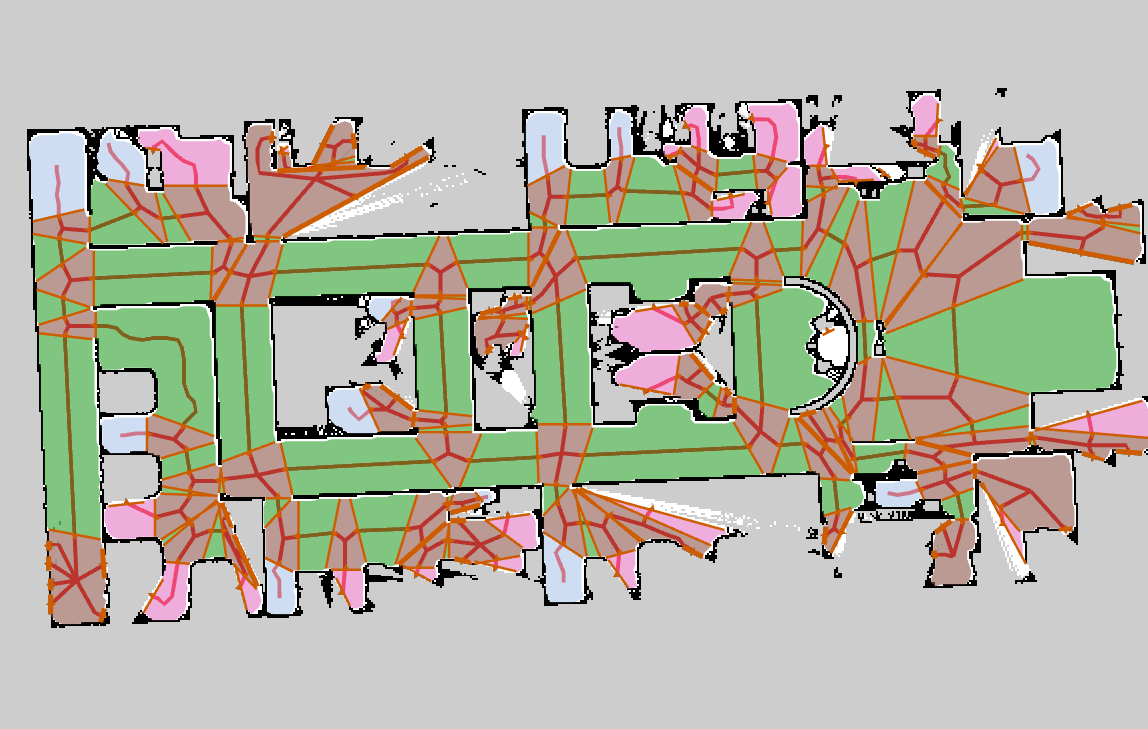}
         \caption{$n_{dir}=5$}
         \label{fig:opt5}
     \end{subfigure}
     \begin{subfigure}[b]{0.24\textwidth}
         \centering
         \includegraphics[width=\textwidth]{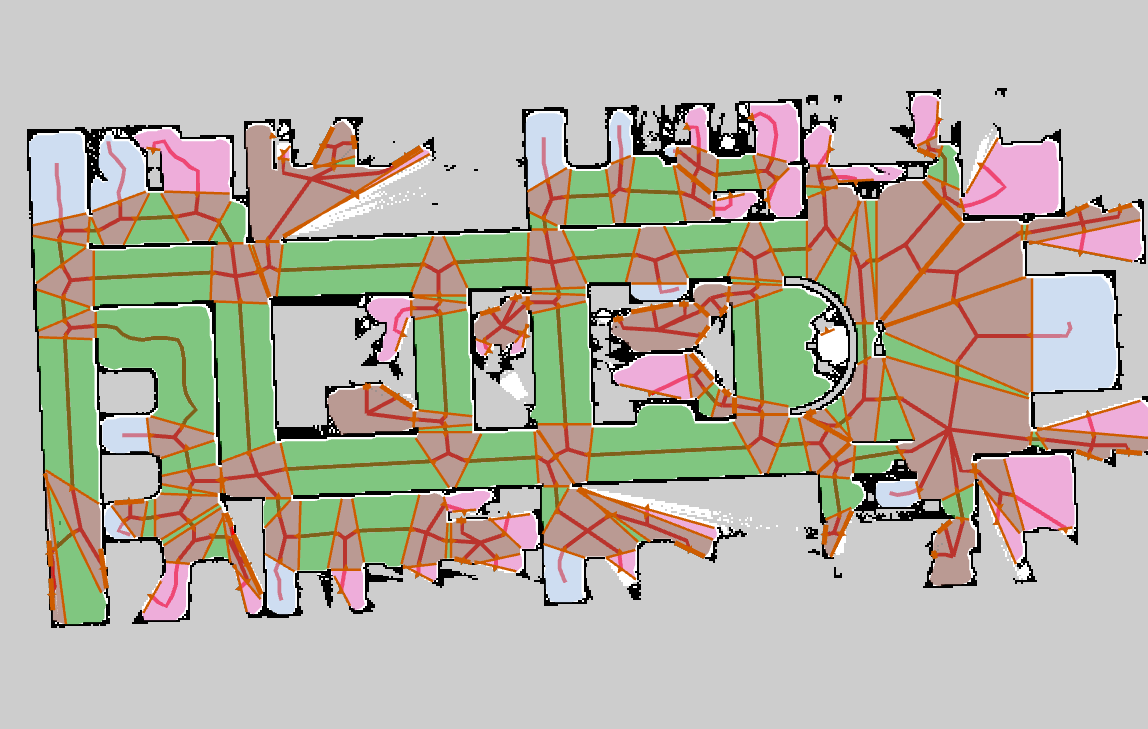}
         \caption{$n_{dir}=6$}
         \label{fig:opt6}
     \end{subfigure}
     \begin{subfigure}[b]{0.24\textwidth}
         \centering
         \includegraphics[width=\textwidth]{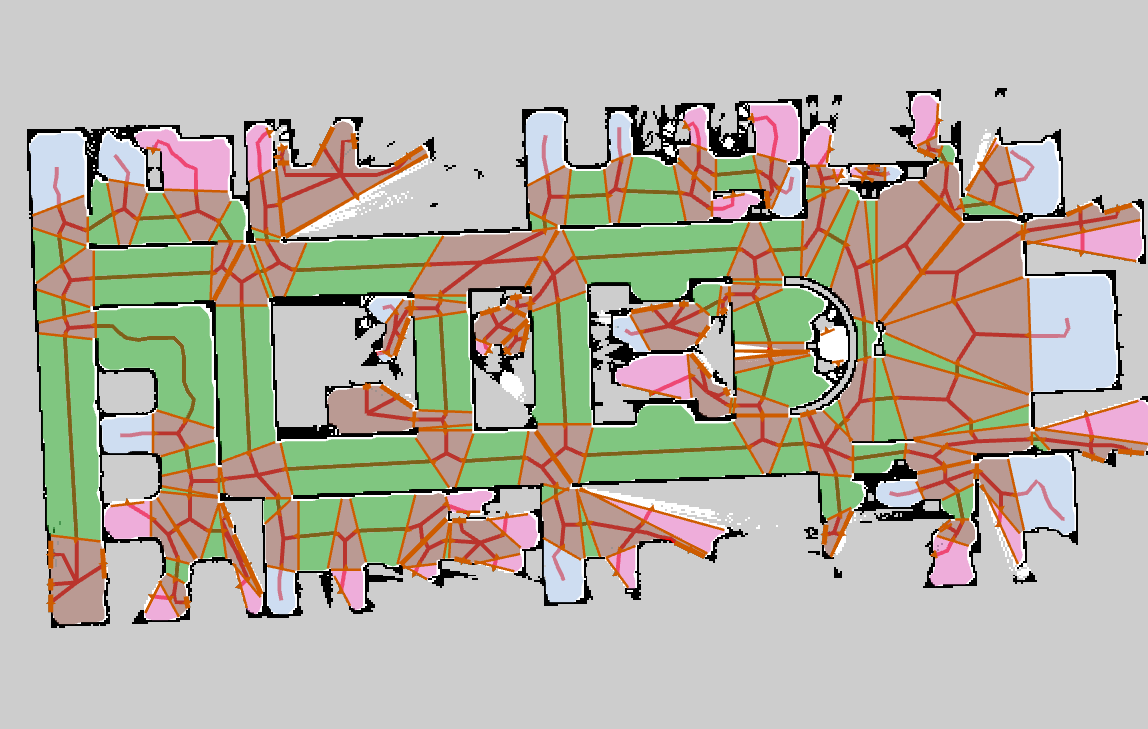}
         \caption{$n_{dir}=7$}
         \label{fig:opt7}
     \end{subfigure}
     \begin{subfigure}[b]{0.24\textwidth}
         \centering
         \includegraphics[width=\textwidth]{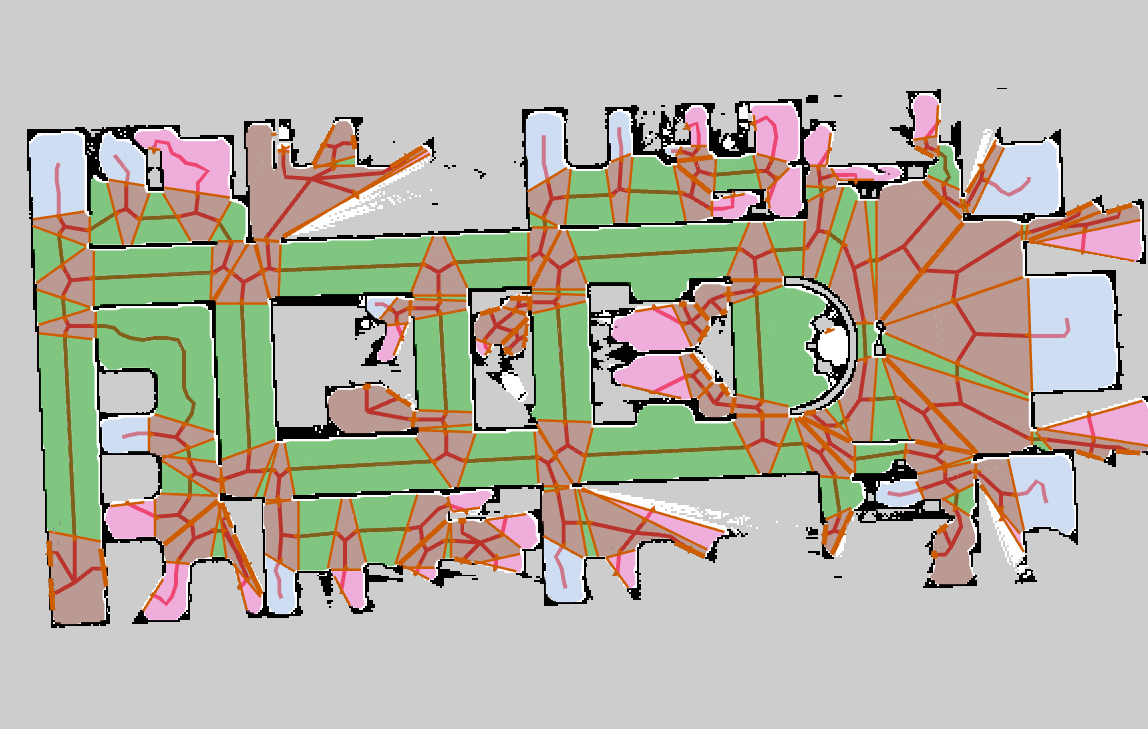}
         \caption{$n_{dir}=8$}
         \label{fig:opt8}
     \end{subfigure}
    \caption{The effect of the choice of the parameter $n_{dir}$ on the output of GRID-FAST semantic and topometric mapping.}
    \label{fig:angles_opt}
\end{figure*}

This subsection describes the process employed in tuning the parameters of the GRID-FAST method. The tuning method is illustrated on the map associated with the first test scenario: the map of a Hospital (Table \ref{tab:maps} and Fig. \ref{fig:angles_opt}). The topometric map will be generated for a robot with a working area of $R_{min}=0.6m$. 

\begin{figure}[H]
    \centering
    \begin{subfigure}[b]{0.49\linewidth}
         \centering
         \includegraphics[width=\textwidth]{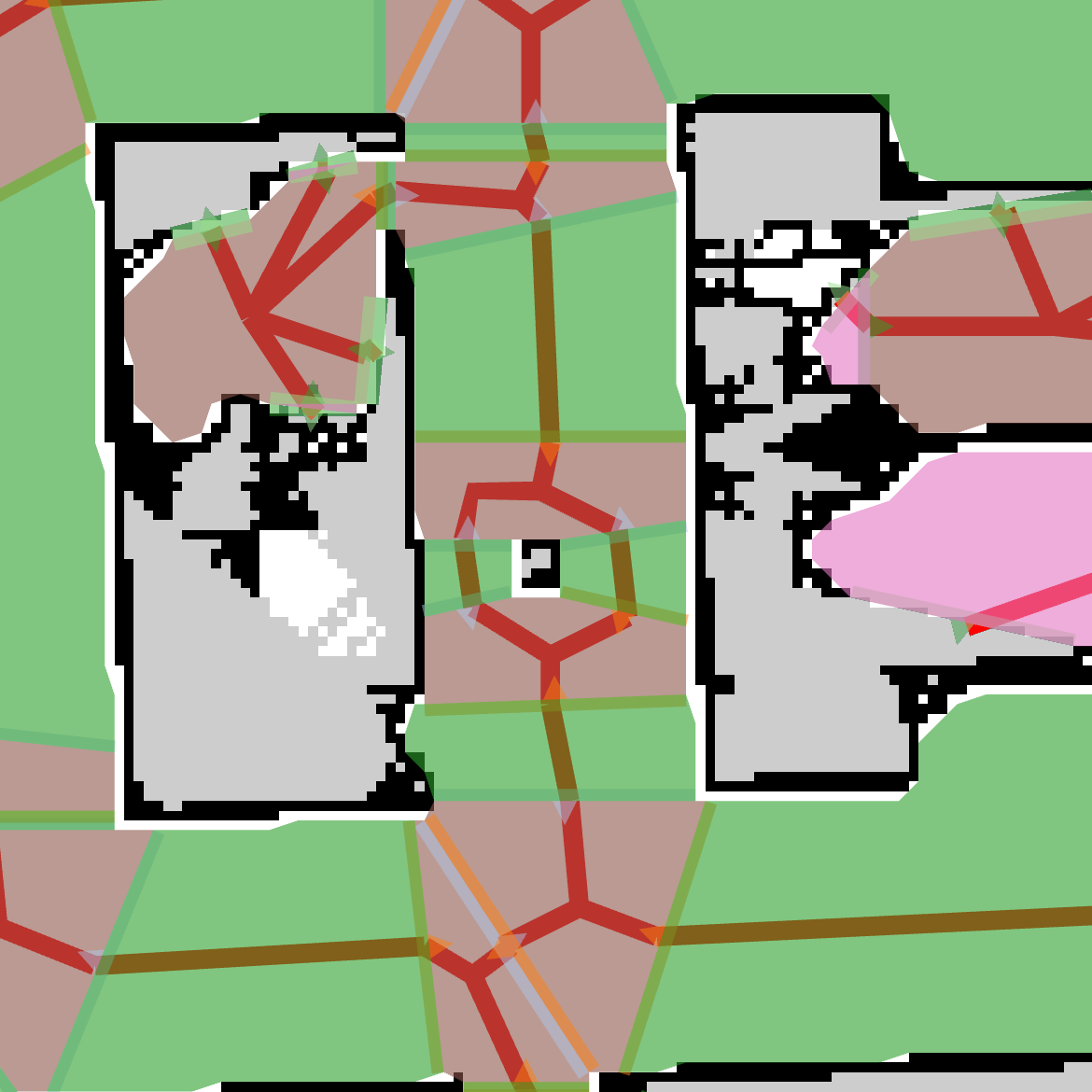}
         \caption{$f_{obj}=10$}
         \label{fig:tuning_object_filter10}
     \end{subfigure}
     \hfill
     \begin{subfigure}[b]{0.49\linewidth}
         \centering
         \includegraphics[width=\textwidth]{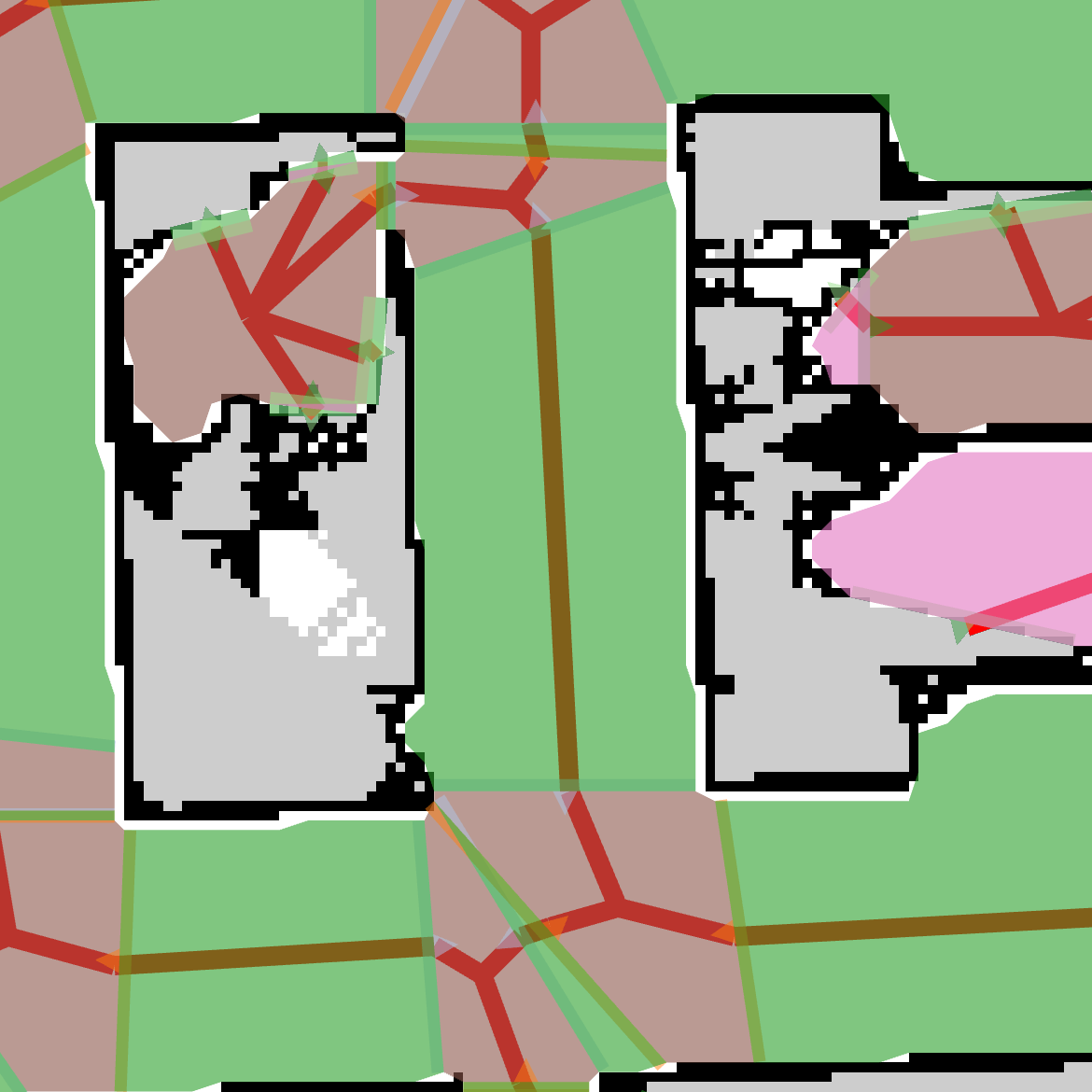}
         \caption{$f_{obj}=20$}
         \label{fig:tuning_object_filter20}
     \end{subfigure}
    \caption{The parameter $f_{obj}$ is tuned such that small objects in the map are removed.}
    \label{fig:tuning_object_filter}
\end{figure}

The parameter $f_{obj}$ from sub-section \ref{sec:walls} is chosen to remove small objects. In one of the corridors in the Hospital map, there is a small obstacle, as shown in Fig. \ref{fig:tuning_object_filter10}. $f_{obj}$ is set to the value 20 to remove the small object, along with all other smaller objects, as shown in Fig. \ref{fig:tuning_object_filter20}. As discussed in Section \ref{sec:walls}, removing objects will result in a simpler topometric map, but at the cost of requiring the robot to have a local navigation system that can avoid obstacles in its path. Since a robot navigating a hospital-like environment with humans and other dynamic objects requires a local navigation system to overcome such unmodelled challenges and is generally equipped with it, it can be justified that the same avoidance system can be used to avoid small objects removed from the map.

\begin{figure}[H]
    \centering
    \begin{subfigure}[b]{0.49\linewidth}
         \centering
         \includegraphics[width=\textwidth]{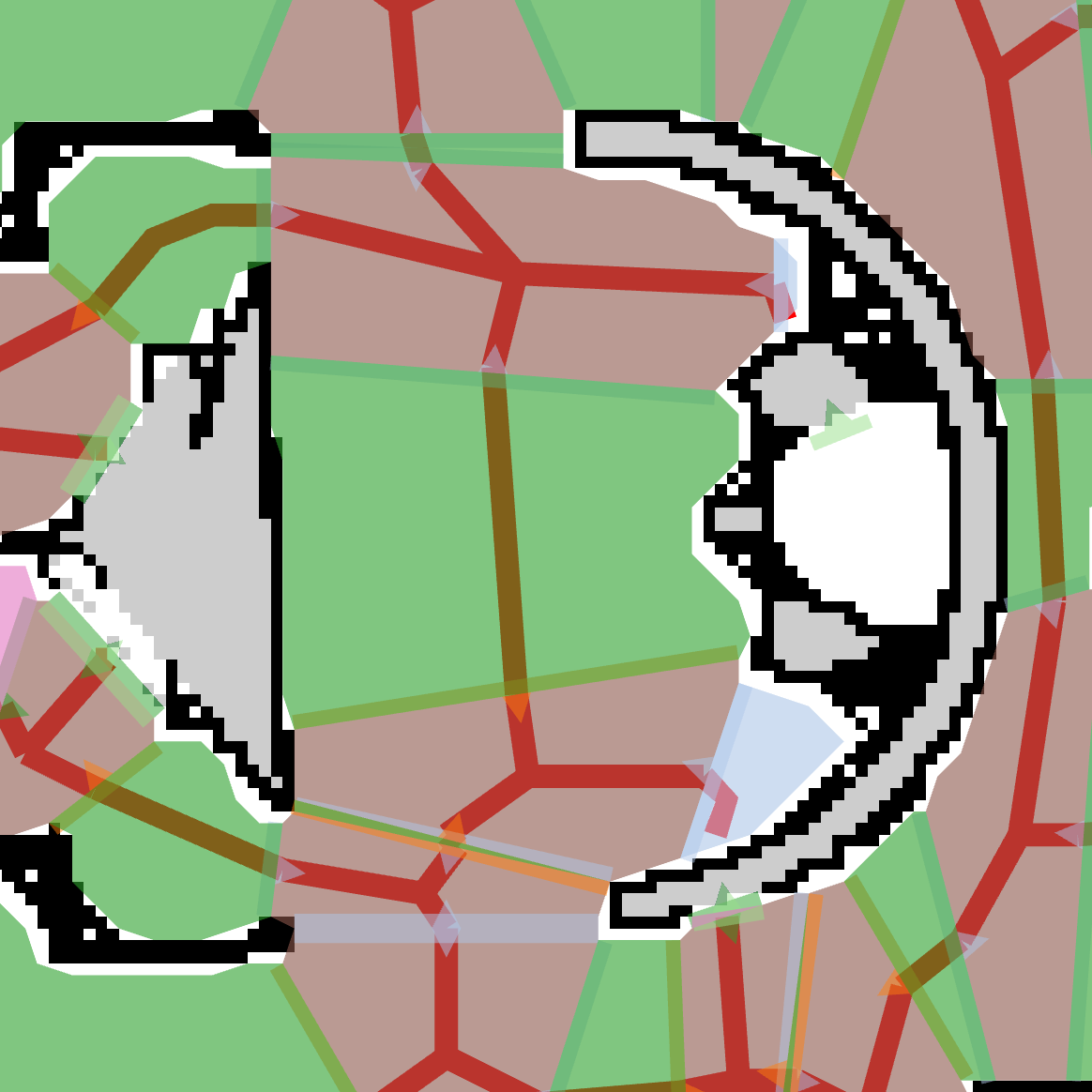}
         \caption{$D_{s}=1$}
         \label{fig:tuning_minDesent1}
     \end{subfigure}
     \hfill
     \begin{subfigure}[b]{0.49\linewidth}
         \centering
         \includegraphics[width=\textwidth]{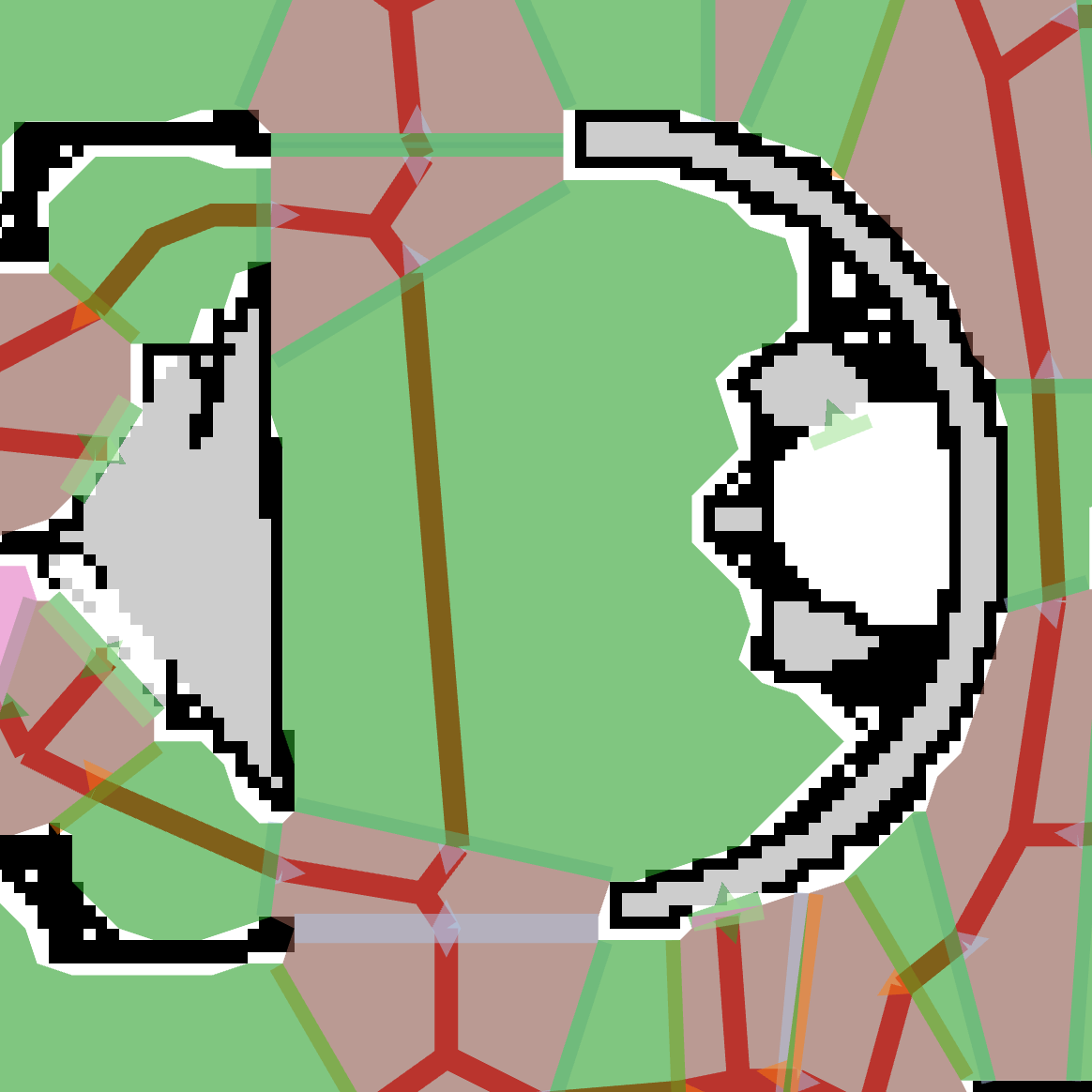}
         \caption{$D_{s}=5$}
         \label{fig:tuning_minDesent5}
     \end{subfigure}
    \caption{The parameter $D_{s}$ is tuned such that the unevenness in the walls is not treated as pathways.}
    \label{fig:tuning_minDesent}
\end{figure}

The following parameter values were chosen for the optimization step in sub-section \ref{sec:interOpt}. The value of $d_{cMin}$ was set to $R_{min}$. A greater or lesser value can be used for $d_{cMin}$, if the robot has worse or better steering performance respectively. The parameter $D_{s}$ was tuned to a value of 5 in the hospital area so that small openings in the reception desk were not detected as openings, as shown in Fig. \ref{fig:tuning_minDesent}. The value $D_{max}$ is set to a large value; in this case, $10^6$ was used.

To ascertain the effect of the value of the parameter $n_{dir}$, GRID-FAST was tested using different values of $n_{dir}$. Fig. \ref{fig:angles_opt} illustrates the output of the GRID-FAST method applied to the Hospital case with various values of $n_{dir}$. 
The value of $n_{dir}$ was selected as 4, as it is the lowest number of $n_{dir}$ required to detect almost all the intersections in the map. 

\begin{table}[h]
\caption{Parameter values used for the GRID-FAST method}
\label{tab:val}
\centering
\begin{tabular}{ll|ll}
\hline
Variable & Value & Variable & Value   \\ \hline
 $R_{min}$& 0.6 m &  $f_{obj}$ & 20 cells \\
 $n_{dir}$ & 4 &  $D_{s}$ & 5 \\
 $f_{uk}$& 3 cell &  $D_{max}$ & $10^6$  \\ \hline
\end{tabular}
\end{table}
%
%------------------------
%
\subsection{Comparison with the method in \cite{fredriksson2023semantic}}\label{Comparison_old_method}

\begin{figure}[ht]
    \centering
    \begin{subfigure}[b]{0.49\linewidth}
         \centering
         \includegraphics[width=\linewidth]{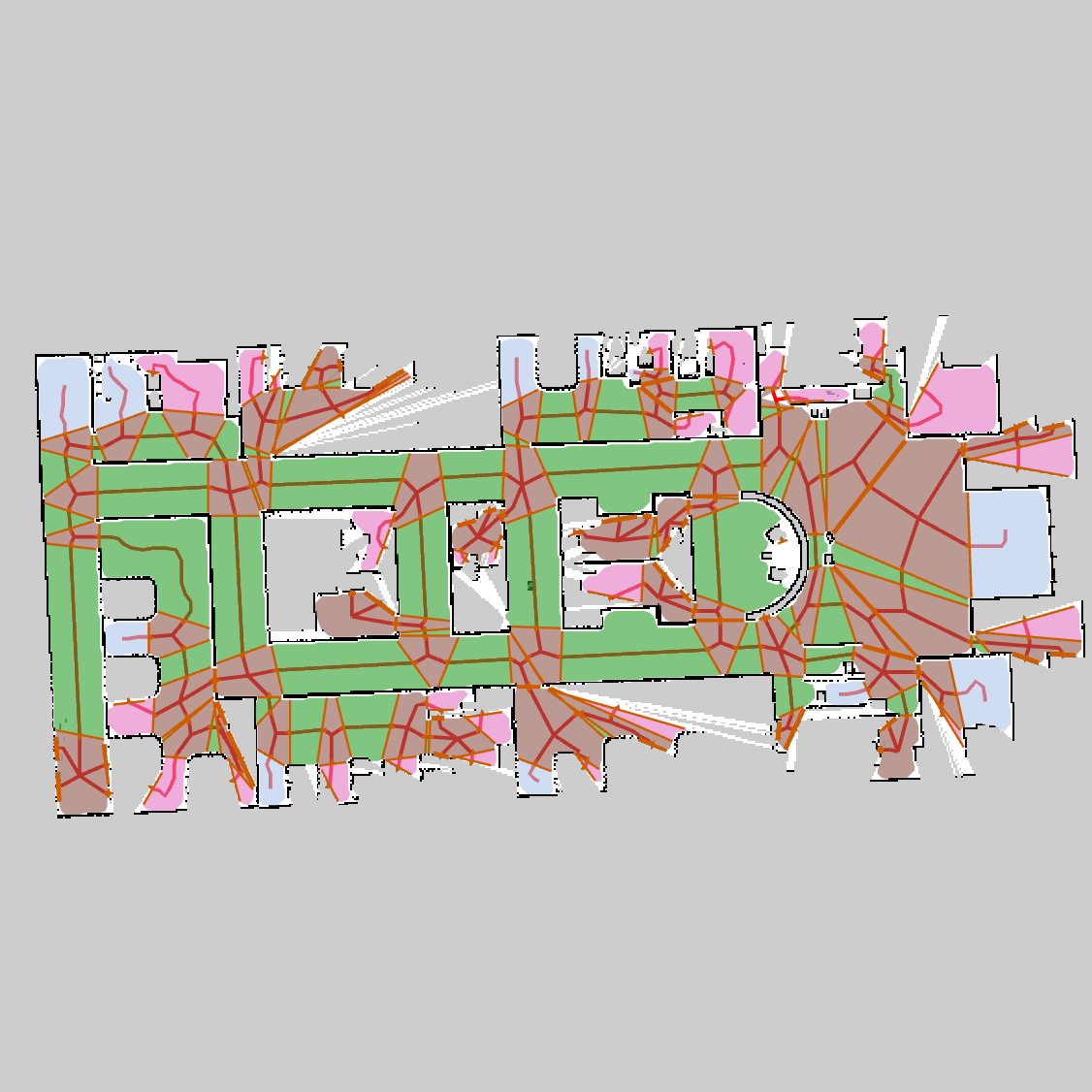}
         \caption{}
         \label{fig:HospitalNew}
     \end{subfigure}
     \hfill
     \begin{subfigure}[b]{0.49\linewidth}
         \centering
         \includegraphics[width=\textwidth]{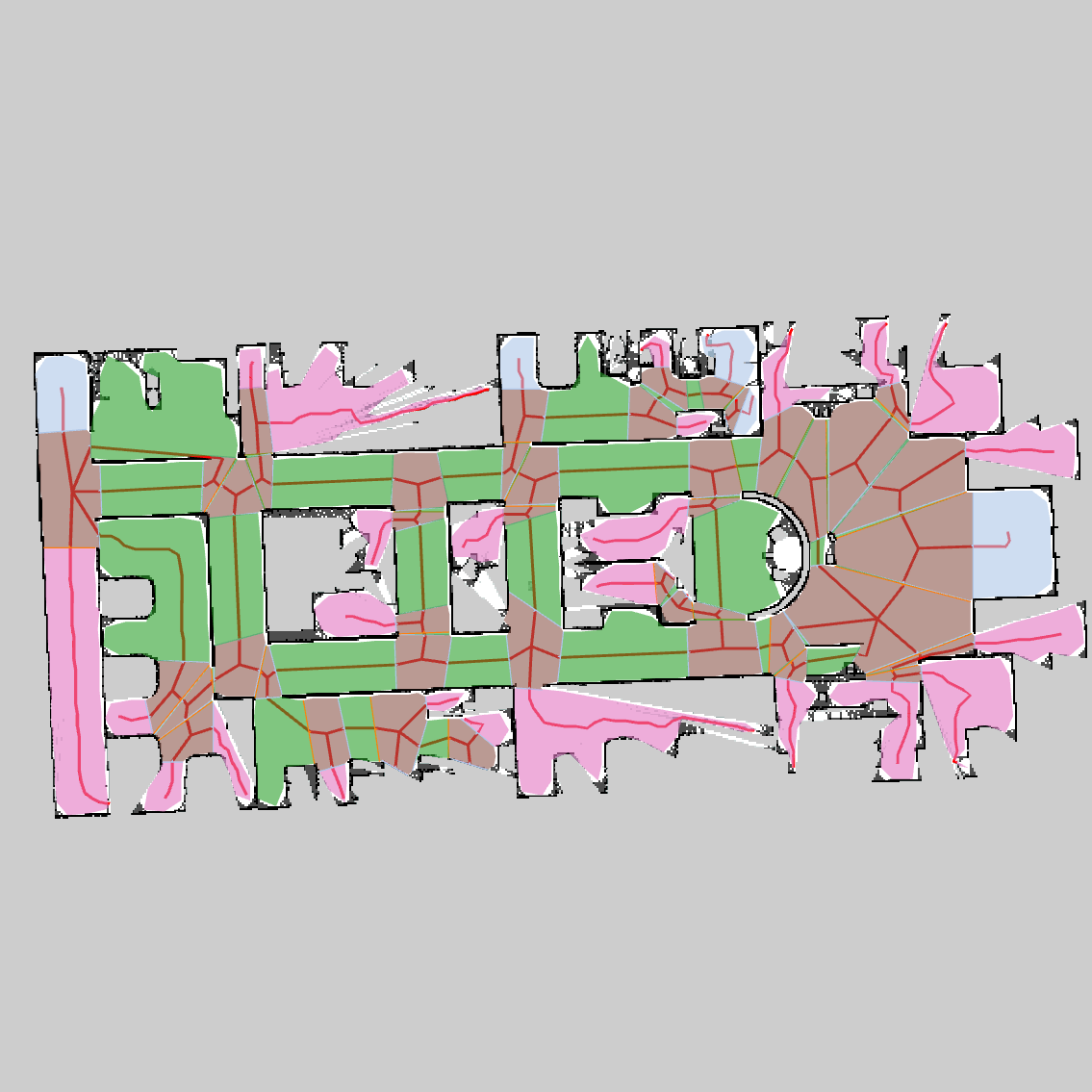}
         \caption{}
         \label{fig:HospitalOld}
     \end{subfigure}
    \caption{Comparison between (a) GRID-FAST and (b) the method presented in \cite{fredriksson2023semantic}. GRID-FAST is able to detect smaller intersections in rooms while requiring fewer scanning directions ($n_{dir}$). GRID-FAST also generates robot paths that are safer than the method in \cite{fredriksson2023semantic}\blue{, as it uses additional safety checks when generating the global paths}.}
    \label{fig:Hospital_old_method}
\end{figure}

The comparison between the preliminary results presented in our earlier work \cite{fredriksson2023semantic} and the results from the enhanced method (GRID-FAST) presented in this article reveal several key advantages of the enhanced method. Firstly, the new method can detect all major intersections that the method in \cite{fredriksson2023semantic} failed to identify. An example of this can be seen in the top left corner of the maps in Figure \ref{fig:Hospital_old_method}. Additionally, the updated method ensured that the generated robot paths did not come too close to walls, thus reducing the potential for collisions; this is something the method in \cite{fredriksson2023semantic} failed within the top left and lower right corners of the map, as can be seen in Figure \ref{fig:HospitalOld}.

The new method required only four scanning angles ($n_{dir}=4$), as opposed to the six scanning angles ($n_{dir}=6$) utilized by the earlier method. This reduction in the number of angles improved the overall computational efficiency of the method, as shown in Table \ref{tab:resOldMethod}. Moreover, the new method exhibited enhanced frontier detection. As a result, the new method performs better in parts of the map that are partially unexplored. Combined with improved intersection detection, this leads to GRID-FAST having more nodes than the method in \cite{fredriksson2023semantic}. \blue{Finally, the intersections created by the new method were smaller and more accurately captured the intersections, thus demonstrating the superiority of the updated approach.}

\begin{table}[h]
\caption{Performance of GRID-FAST and method in \cite{fredriksson2023semantic}.}
\label{tab:resOldMethod}
\centering
%\resizebox{\columnwidth}{!}{
\begin{tabular}{cccc}
\hline
Method & Case &  Number of Nodes & Computation time (ms)\\ \hline
 GRID-FAST  & Hospital &  \blue{133} & \blue{49.9} \\
 \cite{fredriksson2023semantic}  & Hospital &  78 & 61.60  \\  \hline \\
\end{tabular}%}
\end{table}

\subsection{Comparison with State-of-the-art Solutions}
All available solutions for topometric mapping focus on room identification. As GRID-FAST employs intersection detection, its approach has more similarities with Voronoi graphs or related methods. Therefore, in this section, we present a comprehensive comparison of the results of the GRID-FAST approach with available reproducible state-of-the-art Voronoi techniques. 
Only two general approaches are available for generating Voronoi diagrams from a grid-based 2D map. The first approach is from \cite{Beeson}, which employs a thinning method \cite{Zhang1984} to generate extended Voronoi graphs (EVG) \cite{Beeson2005} and reduced general Voronoi graphs (RGVG) \cite{Choset2001}. The second approach is from \cite{Lau2010}, which is used to create general Voronoi graphs (GVG).

\begin{figure}[t]
    \centering
    \begin{subfigure}[b]{0.24\textwidth}
         \centering
         \includegraphics[width=\textwidth]{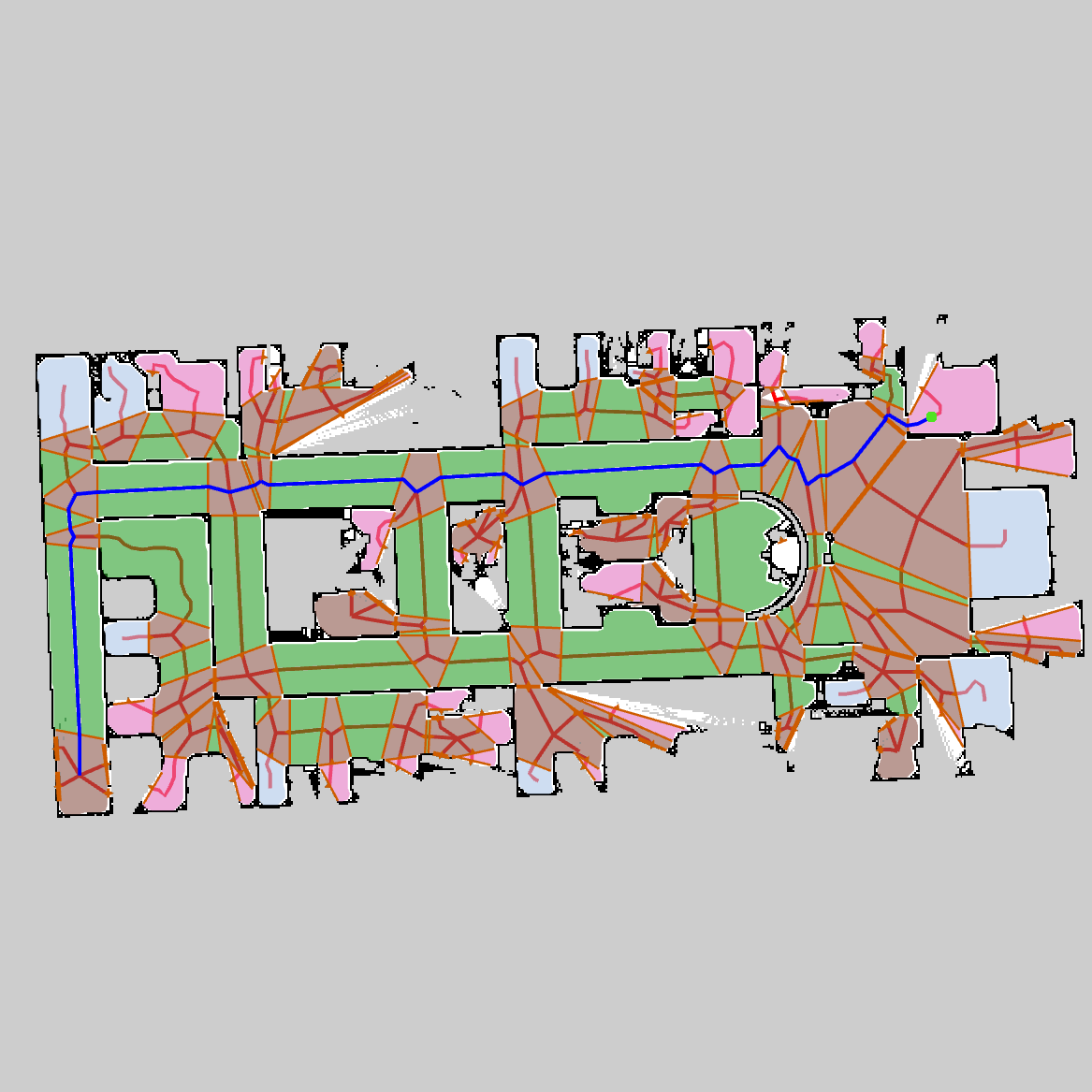}
         \caption{}
         \label{fig:HospitalGRID-FAST}
     \end{subfigure}
    \begin{subfigure}[b]{0.24\textwidth}
         \centering
         \includegraphics[width=\textwidth]{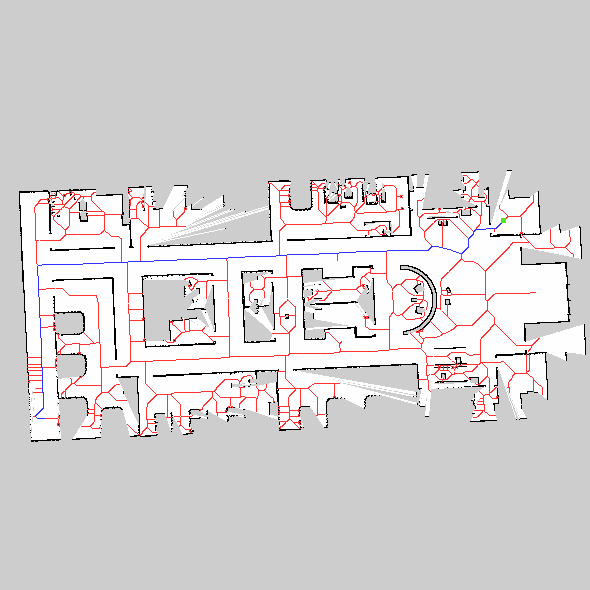}
         \caption{}
         \label{fig:HospitalRGVG}
     \end{subfigure}
     \begin{subfigure}[b]{0.24\textwidth}
         \centering
         \includegraphics[width=\textwidth]{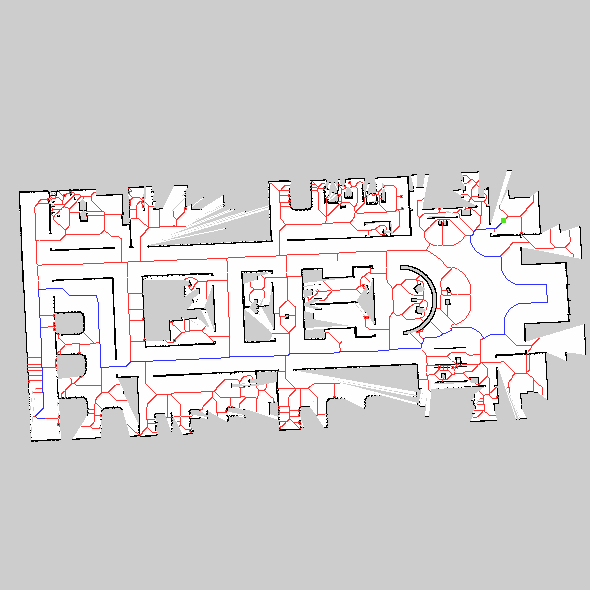}
         \caption{}
         \label{fig:HospitalEVG}
     \end{subfigure}
     \begin{subfigure}[b]{0.24\textwidth}
         \centering
         \includegraphics[width=\textwidth]{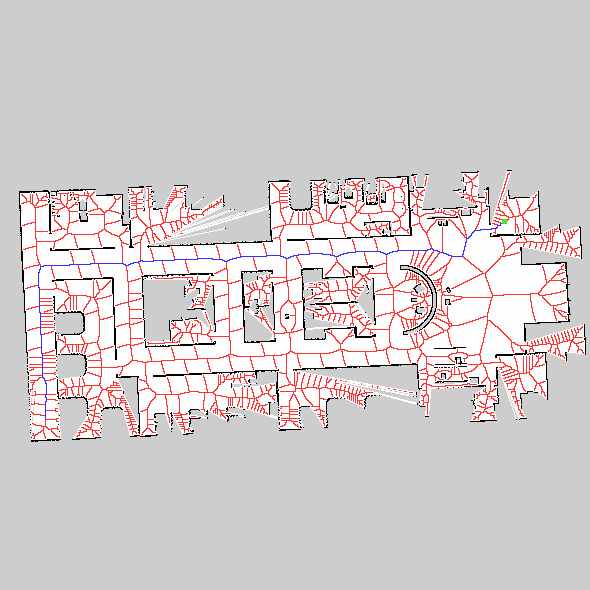}
         \caption{}
         \label{fig:HospitalGVG}
     \end{subfigure}
    \caption{Comparison of robot paths generated with different methods in the hospital map. \blue{The path determined using the A* planner is indicated by the blue line, with the starting point marked by a green marker.} (a) the result of GRID-FAST is shown on the filtered map along with the semantic topometric map. (b) EVG with a sensor horizon of $2m$. (c) RGVG. (d) GVG.}
    \label{fig:Hospital}
\end{figure}

\begin{figure*}[p]
    \centering
    \begin{subfigure}[b]{\textwidth}
         \centering
         \includegraphics[height=.19\textheight]{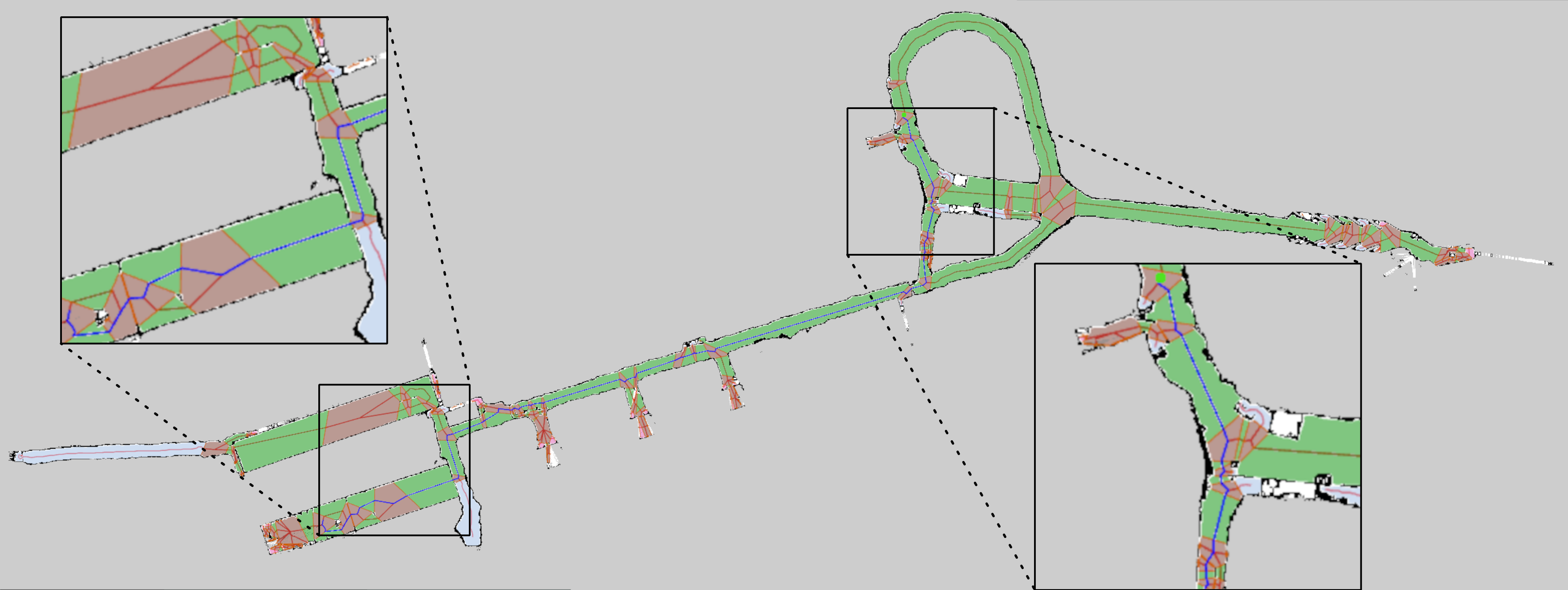}
         \caption{}
         \label{fig:CaveGRID-FAST}
     \end{subfigure}
    \begin{subfigure}[b]{\textwidth}
         \centering
         \includegraphics[height=.19\textheight]{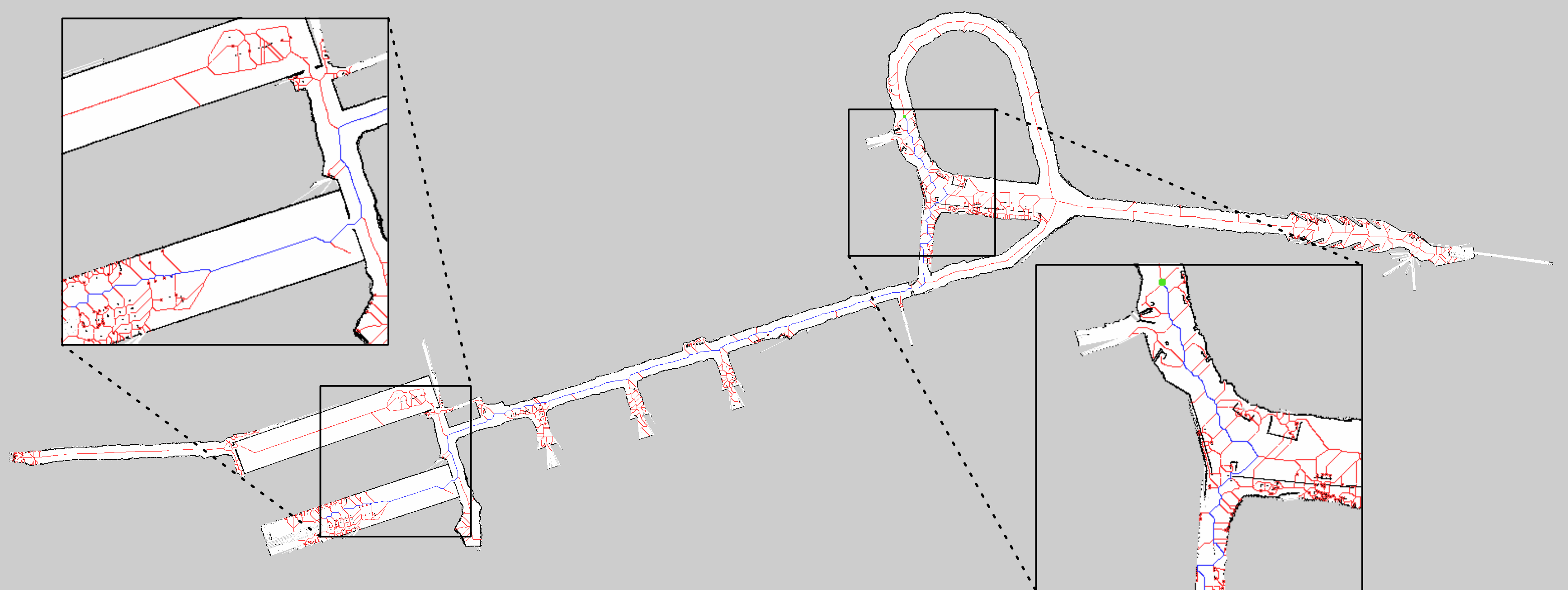}
         \caption{}
         \label{fig:CaveRGVG}
     \end{subfigure}
     \begin{subfigure}[b]{\textwidth}
         \centering
         \includegraphics[height=.19\textheight]{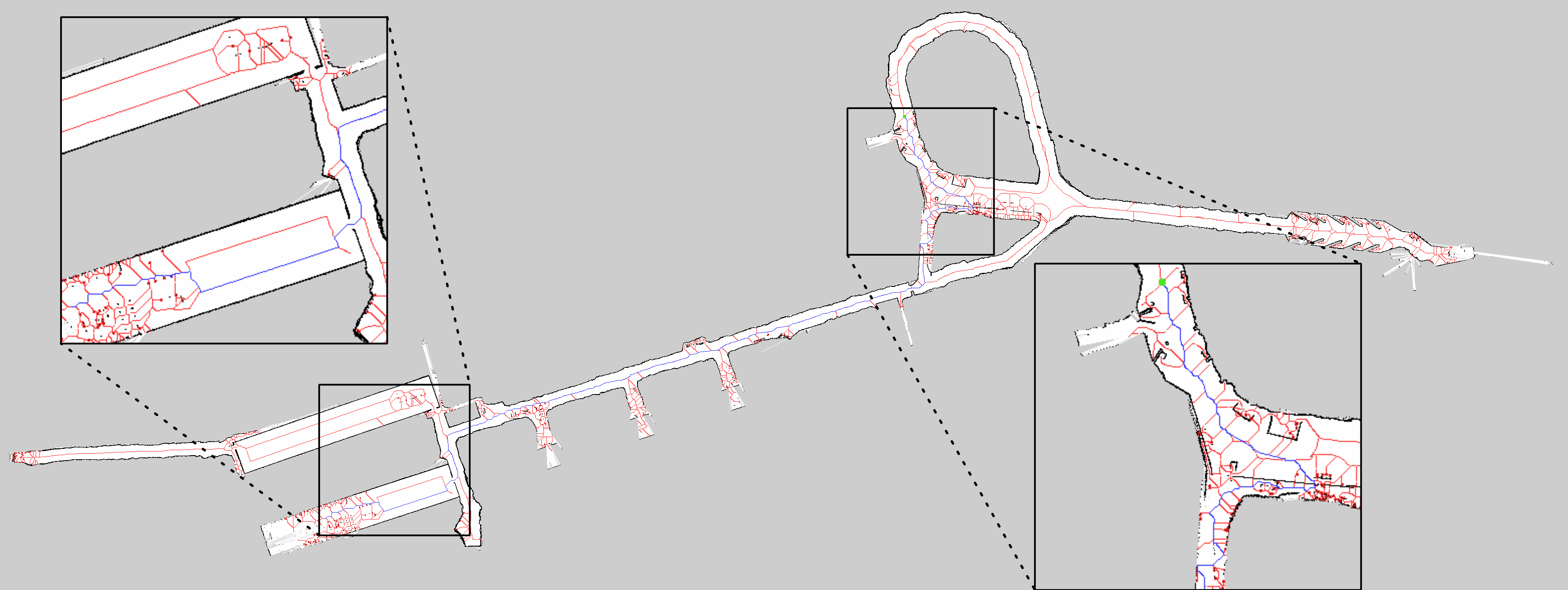}
         \caption{}
         \label{fig:CaveEVG}
     \end{subfigure}
     \begin{subfigure}[b]{\textwidth}
         \centering
        \includegraphics[height=.19\textheight]{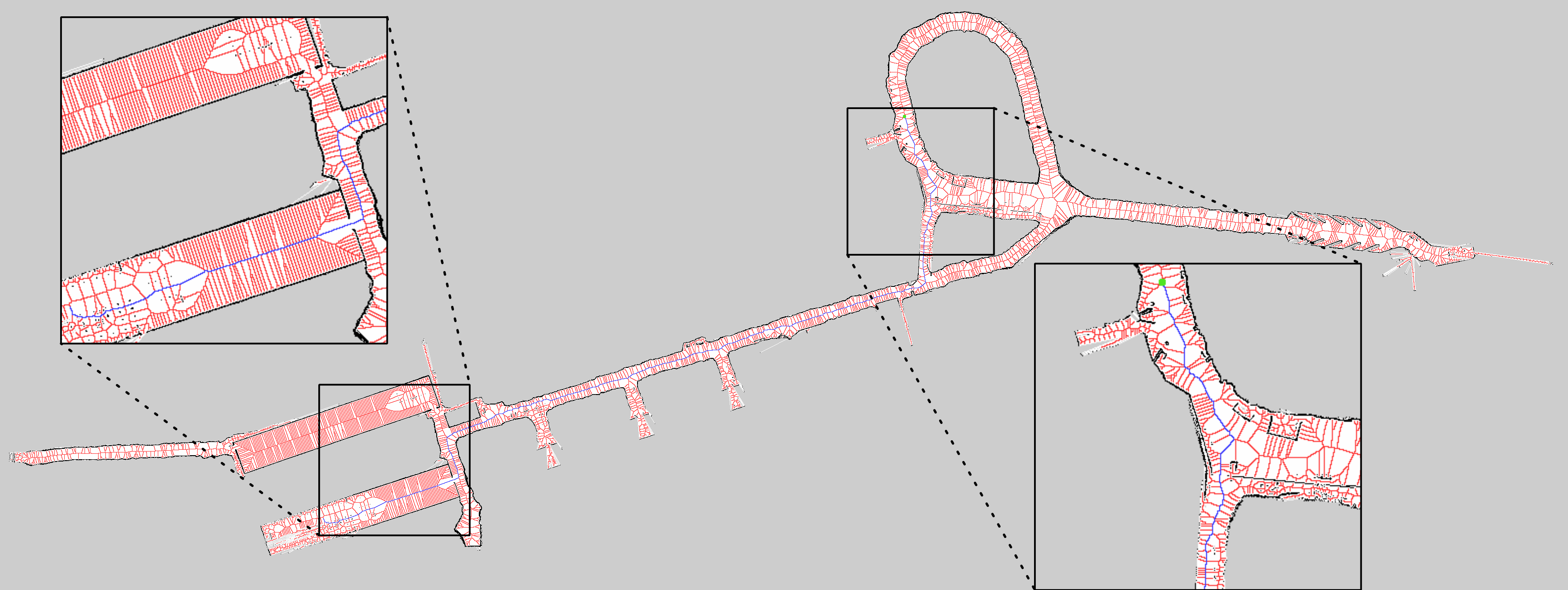}
         \caption{}
         \label{fig:CaveGVG}
     \end{subfigure}
    \caption{Comparison of robot paths generated with different methods in the cave map. \blue{The path determined using the A* planner is indicated by the blue line, with the starting point marked by a green marker.} (a) the result of GRID-FAST is shown on the filtered map along with the semantic topometric map. (b) EVG with a sensor horizon of $2m$. (c) RGVG. (d) GVG.}
    \label{fig:Cave}
\end{figure*}

\begin{figure*}
    \centering
    \begin{subfigure}[b]{.47\textwidth}
         \centering
         \includegraphics[width=\linewidth]{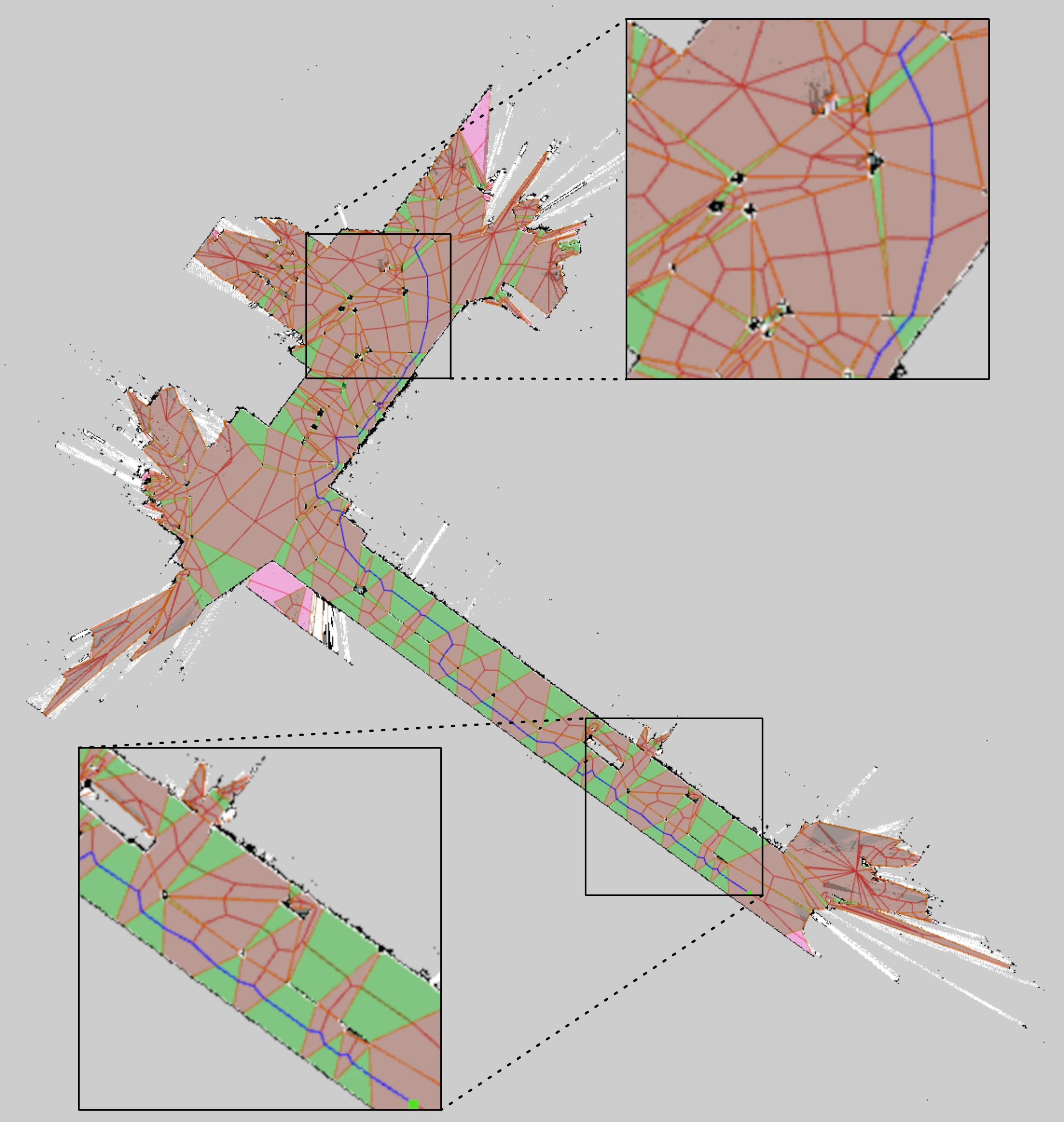}
         \caption{}
         \label{fig:OutdoorGRID-FAST}
     \end{subfigure}
    \begin{subfigure}[b]{.47\textwidth}
         \centering
         \includegraphics[width=\linewidth]{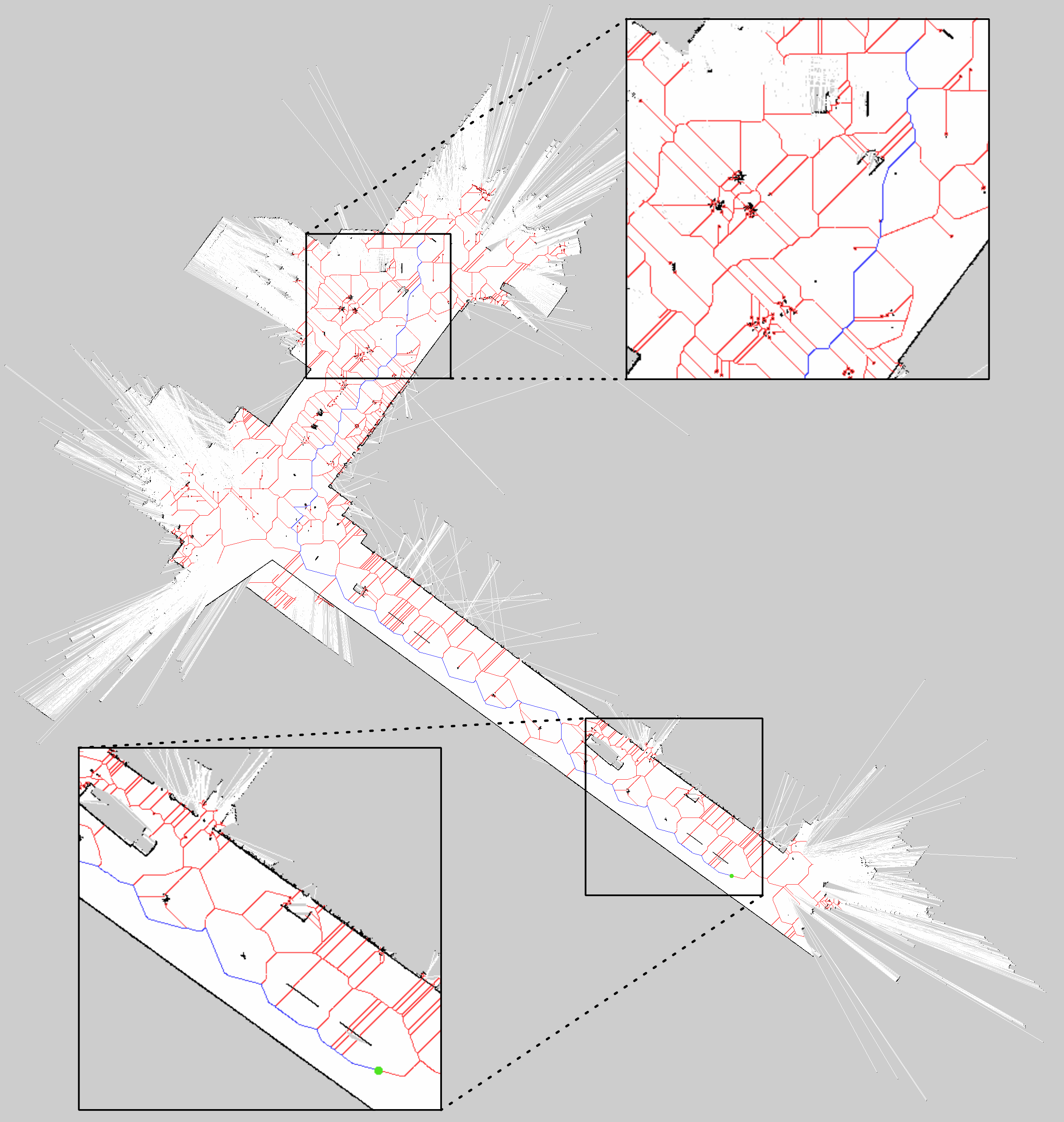}
         \caption{}
         \label{fig:OutdoorRGVG}
     \end{subfigure}
     \begin{subfigure}[b]{.47\textwidth}
         \centering
         \includegraphics[width=\linewidth]{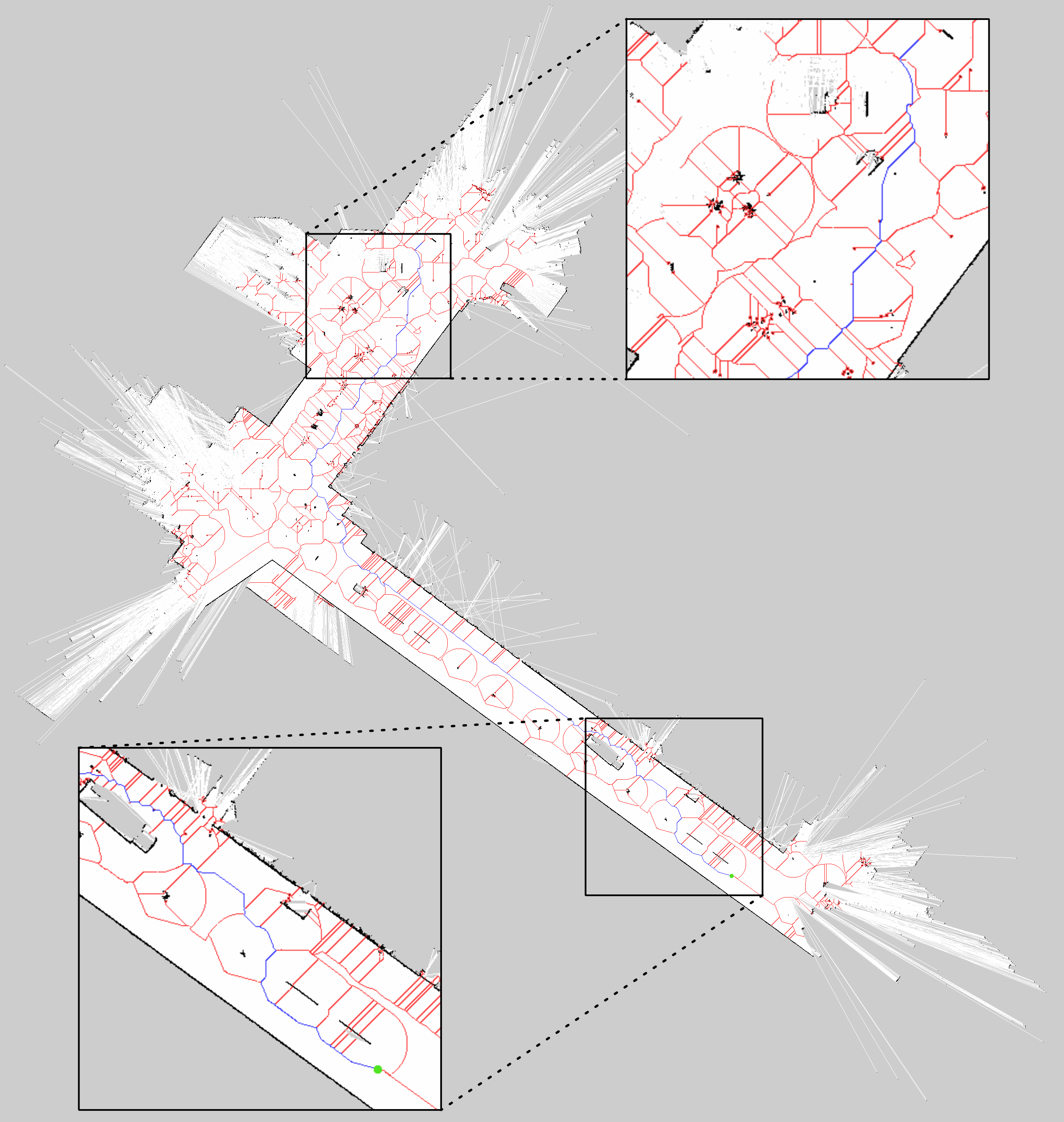}
         \caption{}
         \label{fig:OutdoorEVG}
     \end{subfigure}
     \begin{subfigure}[b]{.47\textwidth}
         \centering
         \includegraphics[width=\linewidth]{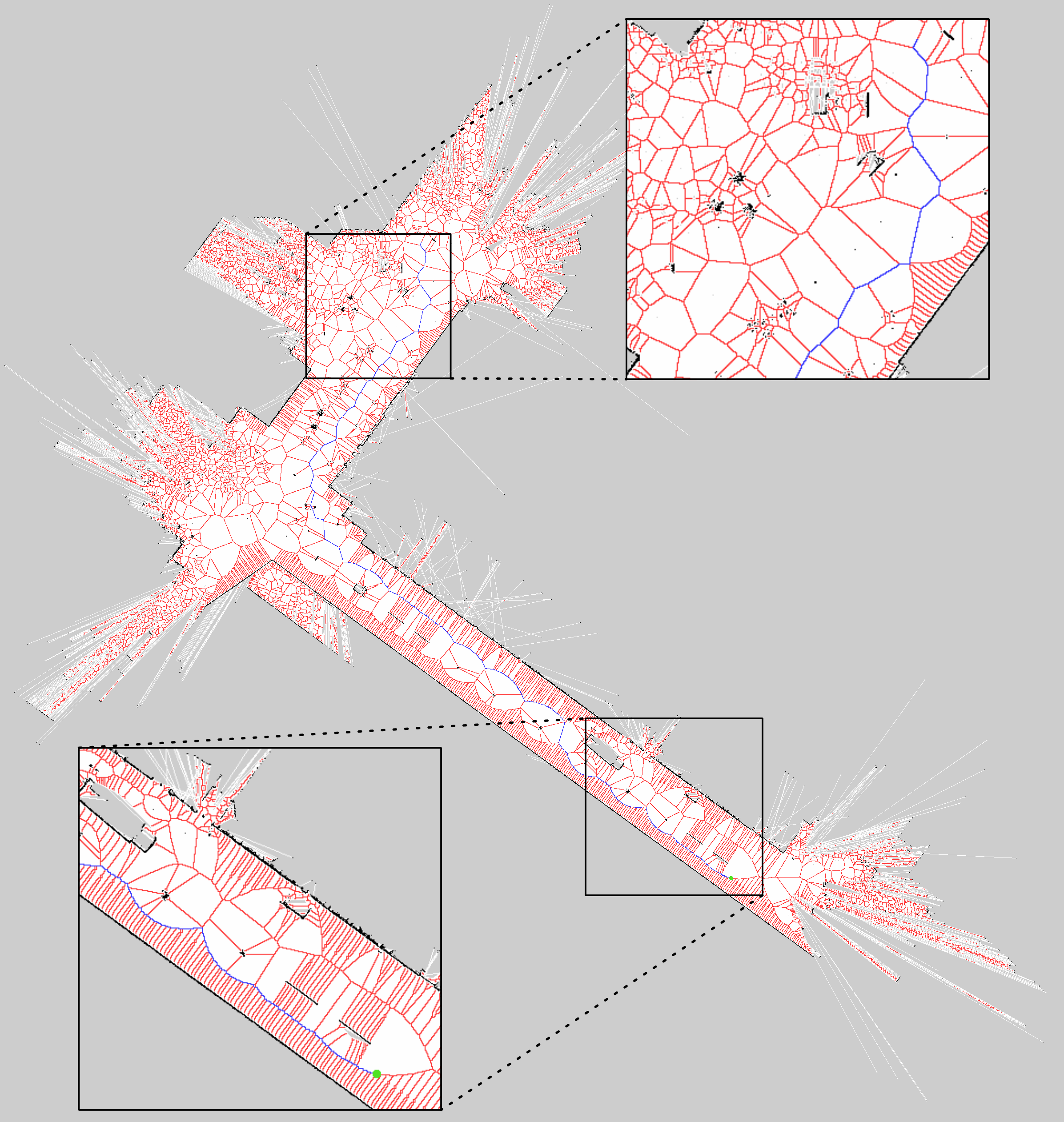}
         \caption{}
         \label{fig:OutdoorGVG}
     \end{subfigure}
    \caption{Comparison of robot paths generated with different methods in the outdoor map. \blue{The path determined using the A* planner is indicated by the blue line, with the starting point marked by a green marker.} (a) the result of GRID-FAST is shown on the filtered map along with the semantic topometric map. (b) EVG with a sensor horizon of $4m$. (c) RGVG. (d) GVG.}
    \label{fig:Outdoor}
\end{figure*}

All the above mentioned methods were tested on three distinct illustrative maps, outlined in Table \ref{tab:maps}. These scenarios vary in size, demonstrating how different methods scale across small, large, and very large maps.
The first map represents a hospital scenario, constructed from LiDAR data collected during the exploration of the hospital gazebo world\footnote{AWS RoboMaker Hospital World ROS package: \\https://github.com/aws-robotics/aws-robomaker-hospital-world}. This case represents an interior environment with rooms, corridors, and intersections.
The second map depicts a cave scenario constructed from LiDAR data from real-world exploration of a subterranean environment \cite{Koval2022}. This map represents subterranean environments such as caves or mines, where intersections are more challenging to define than in the map of an  interior environment.
The third map captures an outdoor scenario constructed from LiDAR data gathered during outdoor exploration on the campus of Luleå University of Technology, Sweden. This scenario evaluates how the solutions handle environments with numerous small obstacles and open spaces.
\begin{table}[htbp]
    \centering
    \caption{The three occupancy maps used in the validation of the GRID-FAST method and comparison with state-of-the-art.}
    \label{tab:maps}
    \begin{tabular}{cccc} \hline
        Map & Resolution & Number of Cells ($10^6$) & Cell Size (m) \\ \hline
        Hospital  &  590x590 & 0.35 & 0.1 \\
        Cave & 3468x1304 & 4.52 & 0.1 \\ 
        Outdoor & 2393x2524 & 6.04 & 0.1\\ \hline
    \end{tabular}
\end{table}
The sensor horizon utilized for the EVG is $2$m, except in the outdoor scenario, where a value of $4$m was employed to create the map successfully. The parameters for GRID-FAST are consistent with those derived in Subsection \ref{sec:tuning} and presented in Table \ref{tab:val}. All maps were generated using Linux, kernel \blue{6.8.6}, running on an AMD 5850u CPU.
\begin{figure}[t]
    \centering
    \includegraphics[width=0.5\linewidth]{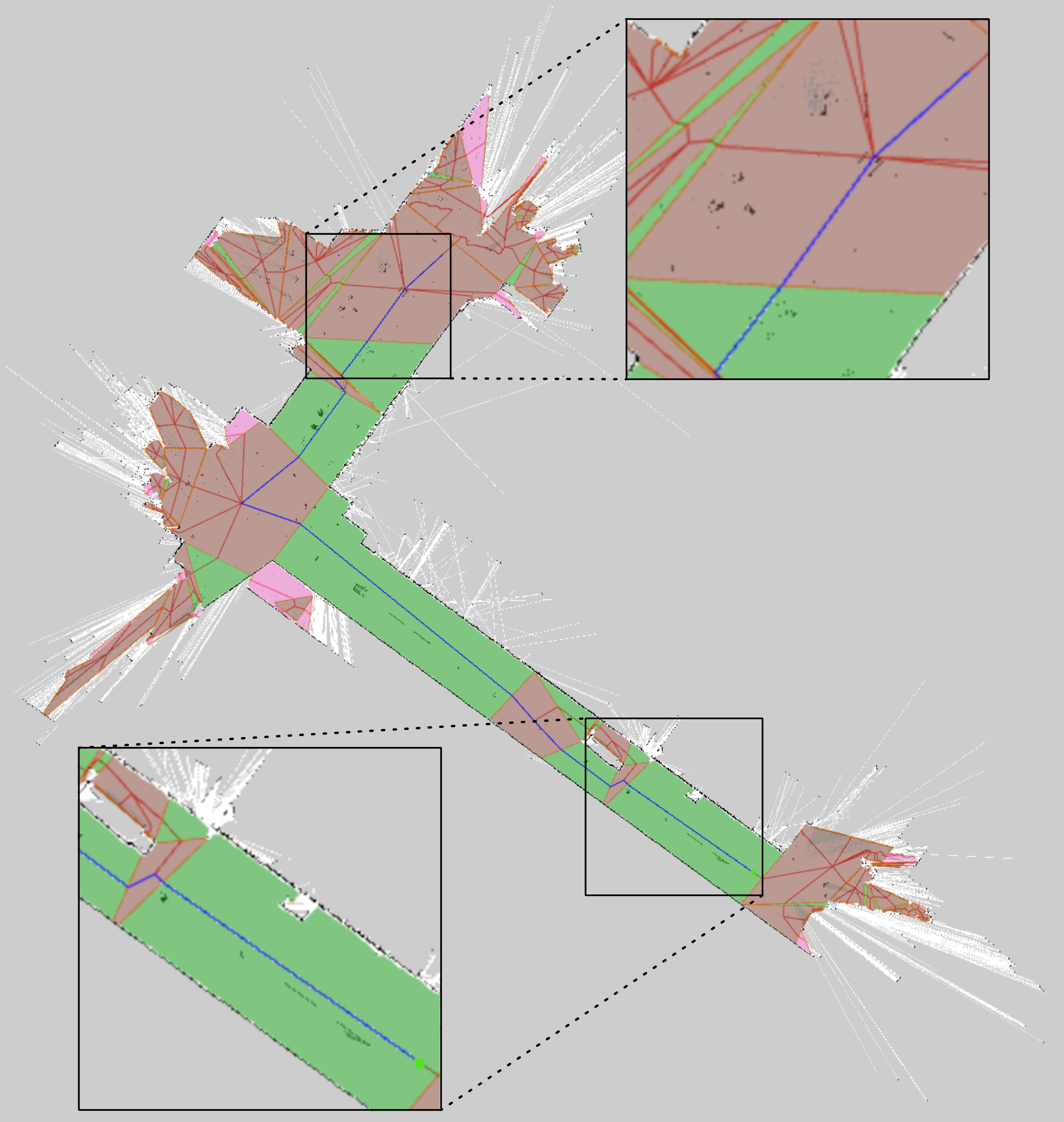}
    \caption{An alternative output of the Outdoor scenario from GRID-FAST using different settings, $f_{obj}=200$ and $R_{min}=1$m. \blue{The path determined using the A* planner is indicated by the blue line, with the starting point marked by a green marker.}}
    \label{fig:OutdoorFilterd}
\end{figure}
\begin{table}[htbp]
\caption{Performance of GRID-FAST and different state-of-the-art topological mapping methods.}
\label{tab:res}
\centering
%\begin{adjustbox}{width=\linewidth}
%\resizebox{.5\textwidth}{!}{%
\begin{tabular}{l|rrrrr}
\hline
&Method &  Nodes & \blue{Graph Length (m)}  & Time (ms) & \blue{Node Searched} \\
&&&&& \blue{(unique nodes)}\\ \hline
 \multirow{4}{*}{\begin{turn}{90}Hospital\end{turn}}
 &GRID-FAST&  \blue{133} & \blue{230.1} & \blue{49.9 (54.6)} & \blue{212(97)} \\
 &RGVG     &  712 & \blue{653.0} & \blue{77.1} & \blue{623(363)}  \\
 %&         &  \blue{[100\%]}&\blue{[100\%]}&\blue{[100\%]}&\blue{[100\%]}\\
 &EVG      &  716 & \blue{656.5} & \blue{37.9} & \blue{1127(651)} \\
 &GVG      &  2823 & \blue{1202.1} & \blue{14.9} & \blue{1637(1006)} \\  \hline
 \multirow{4}{*}{\begin{turn}{90}Cave\end{turn}}
 &GRID-FAST&  \blue{169} & \blue{458.7} & \blue{375.8 (396.5)} & \blue{224(94)}  \\
 &RGVG     &  2131 & \blue{1509.8} & \blue{4011.3} & \blue{2025(983)} \\
 &EVG      &  2405 & \blue{1585.4} & \blue{405.6} & \blue{2562(1236)} \\
 &GVG      &  10640 & \blue{5056.5} & \blue{113.4} & \blue{6363(3746)} \\ \hline 
 
 &GRID-FAST*&  \blue{176} & \blue{867.3} & \blue{632.0 (1247.4)} & \blue{102(45)} \\
 \multirow{3}{*}{\begin{turn}{90}Outdoor \end{turn}}
 &GRID-FAST &  \blue{377} & \blue{1409.8} & \blue{1008.2 (1669.5)} & \blue{381(158)}\\
 &RGVG      &  2089 & \blue{2789.5} & \blue{9840.3} & \blue{2297(1077)} \\
 &EVG       &  2471 & \blue{3020.5} & \blue{1973.8} & \blue{2073(988)} \\
 &GVG       &  21649 & \blue{9600.7} & \blue{210.1} & \blue{6828(3779)} \\
\end{tabular}%
%}
%\end{adjustbox}
\end{table}
\blue{
Four metrics were considered in the comparative study between GRID-FAST and the three Voronoi approaches mentioned previously. 
The first two metrics are the number of nodes, which is the sum of branch points and endpoints, and the total graph length. These metrics indicate the map's sparsity: shorter graph length and fewer nodes suggest a sparser map. 
The third metric is computation time, which measures the time required to generate the topometric and topological maps. For GRID-FAST, computation time is reported in two values: the first is the time taken to generate the topometric map, and the second includes the time taken to generate the topological map.
The final metric involved testing all topological maps in a simulated navigation scenario. An A$^*$ planner \cite{Hart1968} generated a single path between the same two points on all topological maps. The A$^*$ planner's results are provided as two values: the total number of nodes visited and the number of unique nodes visited by the planner.
}
In Fig \ref{fig:Hospital}, \ref{fig:Cave} and, \ref{fig:Outdoor}, we show the skeleton maps generated by the different methods. All topological maps are shown over the original map, except for GRID-FAST, where the topometric map generated by GRID-FAST (with the color scheme shown in Figure \ref{fig:legend}) is overlayed on the filtered version of the original map that is created using methods in subsection \ref{sec:gapDetections} and \ref{sec:walls}.
Table \ref{tab:res} shows the result of the different solutions. The results indicate that GRID-FAST performs at a similar speed to the EVG method but with better scaling for larger maps. This while generating a topological map with significantly fewer nodes \blue{and shorter graph length} compared to all Voronoi-based solutions. When compared to the RGVG method, which ranked second in terms of generating the least number of nodes, GRID-FAST produced, \blue{in worst case 81\% fewer nodes,} in the best case, \blue{$92\%$} fewer nodes and, on average, \blue{$85\%$} fewer nodes. \blue{In the case of the simulated navigation test using the A$^*$ planner, maps produced by GRID-FAST required 65\% fewer nodes visited in the worst case,  89\% fewer nodes visited in the best case and, on average, 79\% fewer nodes compared to the maps produced by RGVG.}
%
%
%\newpage

%\begin{figure*}[!htb]

To highlight another strength of GRID-FAST compared to existing solutions, the method was run two times on the outdoor scenario, where one of the runs (listed in Table \ref{tab:res} (as GRID-FAST*)) use a different value for the parameters $f_{obj}$ and $R_{min}$ than the default value.
%runs used a different set of parameter values
%listed in Table \ref{tab:res} (as GRID-FAST*).
With the default settings, GRID-FAST finds all major intersections on the map. However, as the map contains several small and large scattered objects, the GRID-FAST method yields a topometric and topological map with an excessive number of nodes, which are relatively unimportant in large outdoor scenarios. Considering a robot navigating an outdoor environment, it is more interesting to derive the overall semantic information of the environment. Therefore, the filter value of $f_{obj}$ was increased to $200$ to remove the scattered objects in the environment, and to expand the robot's safety margin, the value of $R_{min}$ was increased to $1$m. The resulting topological map can be seen in Fig. \ref{fig:OutdoorFilterd}.
%
%---------------------------------------------
%
\section{Discussion} \label{sec:discussion}
\subsection{Key difference between GRID-FAST and Voronoi methods}
To generate a solution similar to the one produced by GRID-FAST using Voronoi methods, additional stages must be utilized, such as filtering the map, pruning the topological map, and using the topological map to locate the area and openings of intersections. The strength of GRID-FAST lies not only in proposing a novel approach for a fast method for intersection detection but also in incorporating stages such as filtering and processing of topological maps naturally into the overall algorithm as part of one single solution (captured in Figure \ref{fig:algower}). 
Moreover, from the comparison between GRID-FAST and state-of-the-art Voronoi methods (presented in Table \ref{tab:res}) we see that the entire GRID-FAST procedure is executed 
in a time duration comparable to the time it takes to generate a Voronoi graph in the same environment. This is because the filtering in GRID-FAST utilizes data that the method inherently generates (requiring minimal extra computation time)
and furthermore, GRID-FAST finds intersections without the need to generate a Voronoi graph of the environment, thus saving additional time.   

\subsection{The utility of GRID-FAST for Robotics}
The authors consider the utility of GRID-FAST immense, not solely on its own, but as a foundation for a broader semantic and navigation stack. Due to the sparse nature of the topological map generated by GRID-FAST, it is feasible to find not only a path to a goal location but also the best possible route using navigation rules that would typically be computationally too demanding for conventional navigation solutions \cite{Duchon2014} or less sparse topological map implementations \cite{Lau2010}. In a multi-agent setting, conflict-free routing algorithms can be developed by treating the semantic regions as generalizations of cells in a grid map that require very low memory and computational resources. 

In recent years, methods such as scene graphs \cite{Scene_Graph_ICCV,Chang2023} have emerged, which look to perform robotic tasks based on high-level descriptions. GRID-FAST can become extremely useful in this field as it offers a light and general segmentation of the environment. Moreover, it can easily and seamlessly be appended with more descriptive structural-semantic 
labels generated by other independent solutions, such as door detection, stair detection, and more. 
The semantic objects detected online by the robot (during navigation) can also be added to the topometric map generated by GRID-FAST, creating an all-encompassing semantic map that the robot can use to navigate in its environment intelligently.

\section{Conclusions}\label{sec:conclusion}
This article proposes a novel method for intersection detection, which is then used to generate semantic and topological maps. The approach is general as it is designed with no preconceptions about the map's properties, while taking practical aspects such as the robot's size into account. The distinguishing feature of this approach is that the topometric map is built on intersection detection, making it a more general solution than existing topometric map solutions that use room detection. The topometric map created by GRID-FAST can easily be converted into an extremely sparse topological map in comparison with existing Voronoi-based solutions and can be generated at comparable speeds. 

In the future, we plan to develop an online implementation of the solution 
%in online scenarios 
on hardware robotic platforms and to develop elegant and efficient semantic-based navigation and collision-free routing for a multi-robot system.

%% The Appendices part is started with the command \appendix;
%% appendix sections are then done as normal sections
%% \appendix

%% \section{}
%% \label{}

%% If you have bibdatabase file and want bibtex to generate the
%% bibitems, please use
%%
\bibliography{bliography}

%% BioMed_Central_Bib_Style_v1.01

\begin{thebibliography}{30}
% BibTex style file: bmc-mathphys.bst (version 2.1), 2014-07-24
\ifx \bisbn   \undefined \def \bisbn  #1{ISBN #1}\fi
\ifx \binits  \undefined \def \binits#1{#1}\fi
\ifx \bauthor  \undefined \def \bauthor#1{#1}\fi
\ifx \batitle  \undefined \def \batitle#1{#1}\fi
\ifx \bjtitle  \undefined \def \bjtitle#1{#1}\fi
\ifx \bvolume  \undefined \def \bvolume#1{\textbf{#1}}\fi
\ifx \byear  \undefined \def \byear#1{#1}\fi
\ifx \bissue  \undefined \def \bissue#1{#1}\fi
\ifx \bfpage  \undefined \def \bfpage#1{#1}\fi
\ifx \blpage  \undefined \def \blpage #1{#1}\fi
\ifx \burl  \undefined \def \burl#1{\textsf{#1}}\fi
\ifx \doiurl  \undefined \def \doiurl#1{\url{https://doi.org/#1}}\fi
\ifx \betal  \undefined \def \betal{\textit{et al.}}\fi
\ifx \binstitute  \undefined \def \binstitute#1{#1}\fi
\ifx \binstitutionaled  \undefined \def \binstitutionaled#1{#1}\fi
\ifx \bctitle  \undefined \def \bctitle#1{#1}\fi
\ifx \beditor  \undefined \def \beditor#1{#1}\fi
\ifx \bpublisher  \undefined \def \bpublisher#1{#1}\fi
\ifx \bbtitle  \undefined \def \bbtitle#1{#1}\fi
\ifx \bedition  \undefined \def \bedition#1{#1}\fi
\ifx \bseriesno  \undefined \def \bseriesno#1{#1}\fi
\ifx \blocation  \undefined \def \blocation#1{#1}\fi
\ifx \bsertitle  \undefined \def \bsertitle#1{#1}\fi
\ifx \bsnm \undefined \def \bsnm#1{#1}\fi
\ifx \bsuffix \undefined \def \bsuffix#1{#1}\fi
\ifx \bparticle \undefined \def \bparticle#1{#1}\fi
\ifx \barticle \undefined \def \barticle#1{#1}\fi
\bibcommenthead
\ifx \bconfdate \undefined \def \bconfdate #1{#1}\fi
\ifx \botherref \undefined \def \botherref #1{#1}\fi
\ifx \url \undefined \def \url#1{\textsf{#1}}\fi
\ifx \bchapter \undefined \def \bchapter#1{#1}\fi
\ifx \bbook \undefined \def \bbook#1{#1}\fi
\ifx \bcomment \undefined \def \bcomment#1{#1}\fi
\ifx \oauthor \undefined \def \oauthor#1{#1}\fi
\ifx \citeauthoryear \undefined \def \citeauthoryear#1{#1}\fi
\ifx \endbibitem  \undefined \def \endbibitem {}\fi
\ifx \bconflocation  \undefined \def \bconflocation#1{#1}\fi
\ifx \arxivurl  \undefined \def \arxivurl#1{\textsf{#1}}\fi
\csname PreBibitemsHook\endcsname

%%% 1
\bibitem[\protect\citeauthoryear{Patel et~al.}{2023}]{Patel2023}
\begin{barticle}
\bauthor{\bsnm{Patel}, \binits{A.}},
\bauthor{\bsnm{Lindqvist}, \binits{B.}},
\bauthor{\bsnm{Kanellakis}, \binits{C.}},
\bauthor{\bsnm{Agha-mohammadi}, \binits{A.-a.}},
\bauthor{\bsnm{Nikolakopoulos}, \binits{G.}}:
\batitle{{REF}: {A} {Rapid} {Exploration} {Framework} for {Deploying} {Autonomous} {MAVs} in {Unknown} {Environments}}.
\bjtitle{Journal of Intelligent \& Robotic Systems}
\bvolume{108}(\bissue{3}),
\bfpage{35}
(\byear{2023})
\end{barticle}
\endbibitem

%%% 2
\bibitem[\protect\citeauthoryear{Viswanathan et~al.}{2023}]{Viswanathan2023}
\begin{barticle}
\bauthor{\bsnm{Viswanathan}, \binits{V.K.}},
\bauthor{\bsnm{Lindqvist}, \binits{B.}},
\bauthor{\bsnm{Satpute}, \binits{S.G.}},
\bauthor{\bsnm{Kanellakis}, \binits{C.}},
\bauthor{\bsnm{Nikolakopoulos}, \binits{G.}}:
\batitle{Towards {Visual} {Inspection} of {Distributed} and {Irregular} {Structures}: {A} {Unified} {Autonomy} {Approach}}.
\bjtitle{Journal of Intelligent \& Robotic Systems}
\bvolume{109}(\bissue{2}),
\bfpage{32}
(\byear{2023})
\end{barticle}
\endbibitem

%%% 3
\bibitem[\protect\citeauthoryear{Lindqvist et~al.}{2022}]{Lindqvist2022}
\begin{barticle}
\bauthor{\bsnm{Lindqvist}, \binits{B.}},
\bauthor{\bsnm{Kanellakis}, \binits{C.}},
\bauthor{\bsnm{Mansouri}, \binits{S.S.}},
\bauthor{\bsnm{Agha-mohammadi}, \binits{A.-a.}},
\bauthor{\bsnm{Nikolakopoulos}, \binits{G.}}:
\batitle{{COMPRA}: {A} {COMPact} {Reactive} {Autonomy} {Framework} for {Subterranean} {MAV} {Based} {Search}-{And}-{Rescue} {Operations}}.
\bjtitle{Journal of Intelligent \& Robotic Systems}
\bvolume{105}(\bissue{3}),
\bfpage{49}
(\byear{2022})
\end{barticle}
\endbibitem

%%% 4
\bibitem[\protect\citeauthoryear{Ferguson et~al.}{2005}]{Ferguson2005}
\begin{bchapter}
\bauthor{\bsnm{Ferguson}, \binits{D.}},
\bauthor{\bsnm{Likhachev}, \binits{M.}},
\bauthor{\bsnm{Stentz}, \binits{A.}}:
\bctitle{{A guide to heuristic-based path planning}}.
In: \bbtitle{Proceedings of the International Workshop on Planning Under Uncertainty for Autonomous Systems, International Conference on Automated Planning and Scheduling (ICAPS)},
pp. \bfpage{9}--\blpage{18}
(\byear{2005})
\end{bchapter}
\endbibitem

%%% 5
\bibitem[\protect\citeauthoryear{Remolina and Kuipers}{2004}]{Remolina2004}
\begin{barticle}
\bauthor{\bsnm{Remolina}, \binits{E.}},
\bauthor{\bsnm{Kuipers}, \binits{B.}}:
\batitle{{Towards a general theory of topological maps}}.
\bjtitle{Artificial Intelligence}
\bvolume{152}(\bissue{1}),
\bfpage{47}--\blpage{104}
(\byear{2004})
\end{barticle}
\endbibitem

%%% 6
\bibitem[\protect\citeauthoryear{Kostavelis and Gasteratos}{2015}]{Kostavelis2015}
\begin{barticle}
\bauthor{\bsnm{Kostavelis}, \binits{I.}},
\bauthor{\bsnm{Gasteratos}, \binits{A.}}:
\batitle{{Semantic mapping for mobile robotics tasks: A survey}}.
\bjtitle{Robotics and Autonomous Systems}
\bvolume{66},
\bfpage{86}--\blpage{103}
(\byear{2015})
\end{barticle}
\endbibitem

%%% 7
\bibitem[\protect\citeauthoryear{Bormann et~al.}{2016}]{Bormann2016}
\begin{bchapter}
\bauthor{\bsnm{Bormann}, \binits{R.}},
\bauthor{\bsnm{Jordan}, \binits{F.}},
\bauthor{\bsnm{Li}, \binits{W.}},
\bauthor{\bsnm{Hampp}, \binits{J.}},
\bauthor{\bsnm{Hägele}, \binits{M.}}:
\bctitle{Room segmentation: {Survey}, implementation, and analysis}.
In: \bbtitle{2016 {IEEE} {International} {Conference} on {Robotics} and {Automation} ({ICRA})},
pp. \bfpage{1019}--\blpage{1026}
(\byear{2016})
\end{bchapter}
\endbibitem

%%% 8
\bibitem[\protect\citeauthoryear{Aurenhammer}{1991}]{Aurenhammer1991}
\begin{barticle}
\bauthor{\bsnm{Aurenhammer}, \binits{F.}}:
\batitle{{Voronoi Diagrams—a Survey of a Fundamental Geometric Data Structure}}.
\bjtitle{ACM Comput. Surv.}
\bvolume{23}(\bissue{3}),
\bfpage{345}--\blpage{405}
(\byear{1991})
\end{barticle}
\endbibitem

%%% 9
\bibitem[\protect\citeauthoryear{Mielle et~al.}{2018}]{Mielle2018}
\begin{botherref}
\oauthor{\bsnm{Mielle}, \binits{M.}},
\oauthor{\bsnm{Magnusson}, \binits{M.}},
\oauthor{\bsnm{Lilienthal}, \binits{A.J.}}:
{A method to segment maps from different modalities using free space layout maoris: Map of ripples segmentation}.
Proceedings - IEEE International Conference on Robotics and Automation,
4993--4999
(2018)
\end{botherref}
\endbibitem

%%% 10
\bibitem[\protect\citeauthoryear{Hou et~al.}{2019}]{Hou2019}
\begin{botherref}
\oauthor{\bsnm{Hou}, \binits{J.}},
\oauthor{\bsnm{Yuan}, \binits{Y.}},
\oauthor{\bsnm{Schwertfeger}, \binits{S.}}:
{Area graph: Generation of topological maps using the voronoi diagram}.
2019 19th International Conference on Advanced Robotics, ICAR 2019,
509--515
(2019)
\end{botherref}
\endbibitem

%%% 11
\bibitem[\protect\citeauthoryear{Hiller et~al.}{2019}]{Hiller2019}
\begin{bchapter}
\bauthor{\bsnm{Hiller}, \binits{M.}},
\bauthor{\bsnm{Qiu}, \binits{C.}},
\bauthor{\bsnm{Particke}, \binits{F.}},
\bauthor{\bsnm{Hofmann}, \binits{C.}},
\bauthor{\bsnm{Thielecke}, \binits{J.}}:
\bctitle{{Learning Topometric Semantic Maps from Occupancy Grids}}.
In: \bbtitle{2019 IEEE/RSJ International Conference on Intelligent Robots and Systems (IROS)},
pp. \bfpage{4190}--\blpage{4197}
(\byear{2019})
\end{bchapter}
\endbibitem

%%% 12
\bibitem[\protect\citeauthoryear{He et~al.}{2021}]{He2021}
\begin{barticle}
\bauthor{\bsnm{He}, \binits{Z.}},
\bauthor{\bsnm{Sun}, \binits{H.}},
\bauthor{\bsnm{Hou}, \binits{J.}},
\bauthor{\bsnm{Ha}, \binits{Y.}},
\bauthor{\bsnm{Schwertfeger}, \binits{S.}}:
\batitle{{Hierarchical topometric representation of 3D robotic maps}}.
\bjtitle{Autonomous Robots}
\bvolume{45}(\bissue{5}),
\bfpage{755}--\blpage{771}
(\byear{2021})
\end{barticle}
\endbibitem

%%% 13
\bibitem[\protect\citeauthoryear{Luperto et~al.}{2022}]{Luperto2022}
\begin{barticle}
\bauthor{\bsnm{Luperto}, \binits{M.}},
\bauthor{\bsnm{Kucner}, \binits{T.P.}},
\bauthor{\bsnm{Tassi}, \binits{A.}},
\bauthor{\bsnm{Magnusson}, \binits{M.}},
\bauthor{\bsnm{Amigoni}, \binits{F.}}:
\batitle{Robust {Structure} {Identification} and {Room} {Segmentation} of {Cluttered} {Indoor} {Environments} {From} {Occupancy} {Grid} {Maps}}.
\bjtitle{IEEE Robotics and Automation Letters}
\bvolume{7}(\bissue{3}),
\bfpage{7974}--\blpage{7981}
(\byear{2022})
\end{barticle}
\endbibitem

%%% 14
\bibitem[\protect\citeauthoryear{Blochliger et~al.}{2018}]{Blochliger2018}
\begin{botherref}
\oauthor{\bsnm{Blochliger}, \binits{F.}},
\oauthor{\bsnm{Fehr}, \binits{M.}},
\oauthor{\bsnm{Dymczyk}, \binits{M.}},
\oauthor{\bsnm{Schneider}, \binits{T.}},
\oauthor{\bsnm{Siegwart}, \binits{R.}}:
{Topomap: Topological Mapping and Navigation Based on Visual SLAM Maps}.
Proceedings - IEEE International Conference on Robotics and Automation,
3818--3825
(2018)
\end{botherref}
\endbibitem

%%% 15
\bibitem[\protect\citeauthoryear{Park et~al.}{2018}]{Park2018}
\begin{barticle}
\bauthor{\bsnm{Park}, \binits{B.}},
\bauthor{\bsnm{Choi}, \binits{J.}},
\bauthor{\bsnm{Chung}, \binits{W.K.}}:
\batitle{{Incremental hierarchical roadmap construction for efficient path planning}}.
\bjtitle{ETRI Journal}
\bvolume{40}(\bissue{4}),
\bfpage{458}--\blpage{470}
(\byear{2018})
\end{barticle}
\endbibitem

%%% 16
\bibitem[\protect\citeauthoryear{Kwon and Song}{2008}]{Kwon2008}
\begin{barticle}
\bauthor{\bsnm{Kwon}, \binits{T.B.}},
\bauthor{\bsnm{Song}, \binits{J.B.}}:
\batitle{{Real-time building of a thinning-based topological map}}.
\bjtitle{Intelligent Service Robotics}
\bvolume{1}(\bissue{3}),
\bfpage{211}--\blpage{220}
(\byear{2008})
\end{barticle}
\endbibitem

%%% 17
\bibitem[\protect\citeauthoryear{Lau et~al.}{2010}]{Lau2010}
\begin{bchapter}
\bauthor{\bsnm{Lau}, \binits{B.}},
\bauthor{\bsnm{Sprunk}, \binits{C.}},
\bauthor{\bsnm{Burgard}, \binits{W.}}:
\bctitle{{Improved updating of Euclidean distance maps and Voronoi diagrams}}.
In: \bbtitle{2010 IEEE/RSJ International Conference on Intelligent Robots and Systems},
pp. \bfpage{281}--\blpage{286}
(\byear{2010})
\end{bchapter}
\endbibitem

%%% 18
\bibitem[\protect\citeauthoryear{Choset and Burdick}{2000}]{Choset2000}
\begin{barticle}
\bauthor{\bsnm{Choset}, \binits{H.}},
\bauthor{\bsnm{Burdick}, \binits{J.}}:
\batitle{{Sensor-Based Exploration: The Hierarchical Generalized Voronoi Graph}}.
\bjtitle{The International Journal of Robotics Research}
\bvolume{19}(\bissue{2}),
\bfpage{96}--\blpage{125}
(\byear{2000})
\end{barticle}
\endbibitem

%%% 19
\bibitem[\protect\citeauthoryear{Choset and Nagatani}{2001}]{Choset2001}
\begin{barticle}
\bauthor{\bsnm{Choset}, \binits{H.}},
\bauthor{\bsnm{Nagatani}, \binits{K.}}:
\batitle{{Topological simultaneous localization and mapping (SLAM): toward exact localization without explicit localization}}.
\bjtitle{IEEE Transactions on Robotics and Automation}
\bvolume{17}(\bissue{2}),
\bfpage{125}--\blpage{137}
(\byear{2001})
\end{barticle}
\endbibitem

%%% 20
\bibitem[\protect\citeauthoryear{Beeson et~al.}{2005}]{Beeson2005}
\begin{bchapter}
\bauthor{\bsnm{Beeson}, \binits{P.}},
\bauthor{\bsnm{Jong}, \binits{N.K.}},
\bauthor{\bsnm{Kuipers}, \binits{B.}}:
\bctitle{{Towards Autonomous Topological Place Detection Using the Extended Voronoi Graph}}.
In: \bbtitle{Proceedings of the 2005 IEEE International Conference on Robotics and Automation},
pp. \bfpage{4373}--\blpage{4379}
(\byear{2005})
\end{bchapter}
\endbibitem

%%% 21
\bibitem[\protect\citeauthoryear{Hou et~al.}{2021}]{Hou2021}
\begin{bchapter}
\bauthor{\bsnm{Hou}, \binits{Q.}},
\bauthor{\bsnm{Zhang}, \binits{S.}},
\bauthor{\bsnm{Chen}, \binits{S.}},
\bauthor{\bsnm{Nan}, \binits{Z.}},
\bauthor{\bsnm{Zheng}, \binits{N.}}:
\bctitle{{Straight Skeleton Based Automatic Generation of Hierarchical Topological Map in Indoor Environment}}.
In: \bbtitle{2021 IEEE International Intelligent Transportation Systems Conference (ITSC)},
pp. \bfpage{2229}--\blpage{2236}
(\byear{2021})
\end{bchapter}
\endbibitem

%%% 22
\bibitem[\protect\citeauthoryear{Fredriksson et~al.}{2023}]{fredriksson2023semantic}
\begin{barticle}
\bauthor{\bsnm{Fredriksson}, \binits{S.}},
\bauthor{\bsnm{Saradagi}, \binits{A.}},
\bauthor{\bsnm{Nikolakopoulos}, \binits{G.}}:
\batitle{Semantic and topological mapping using intersection identification}.
\bjtitle{IFAC-PapersOnLine}
\bvolume{56}(\bissue{2}),
\bfpage{9251}--\blpage{9256}
(\byear{2023}).
\bcomment{22nd IFAC World Congress}
\end{barticle}
\endbibitem

%%% 23
\bibitem[\protect\citeauthoryear{Dunlavey}{1983}]{Dunlavey1983}
\begin{barticle}
\bauthor{\bsnm{Dunlavey}, \binits{M.R.}}:
\batitle{{Efficient Polygon-Filling Algorithms for Raster Displays}}.
\bjtitle{ACM Trans. Graph.}
\bvolume{2}(\bissue{4}),
\bfpage{264}--\blpage{273}
(\byear{1983})
\end{barticle}
\endbibitem

%%% 24
\bibitem[\protect\citeauthoryear{Zhang and Suen}{1984}]{Zhang1984}
\begin{barticle}
\bauthor{\bsnm{Zhang}, \binits{T.Y.}},
\bauthor{\bsnm{Suen}, \binits{C.Y.}}:
\batitle{{A Fast Parallel Algorithm for Thinning Digital Patterns}}.
\bjtitle{Commun. ACM}
\bvolume{27}(\bissue{3}),
\bfpage{236}--\blpage{239}
(\byear{1984})
\end{barticle}
\endbibitem

%%% 25
\bibitem[\protect\citeauthoryear{Beeson}{2006}]{Beeson}
\begin{botherref}
\oauthor{\bsnm{Beeson}, \binits{P.}}:
{EVG-Thin: A Thinning Approximation to the Extended Voronoi Graph}
(2006)
\end{botherref}
\endbibitem

%%% 26
\bibitem[\protect\citeauthoryear{Koval et~al.}{2022}]{Koval2022}
\begin{barticle}
\bauthor{\bsnm{Koval}, \binits{A.}},
\bauthor{\bsnm{Karlsson}, \binits{S.}},
\bauthor{\bsnm{Mansouri}, \binits{S.S.}},
\bauthor{\bsnm{Kanellakis}, \binits{C.}},
\bauthor{\bsnm{Tevetzidis}, \binits{I.}},
\bauthor{\bsnm{Haluska}, \binits{J.}},
\bauthor{\bsnm{Agha-mohammadi}, \binits{A.-a.}},
\bauthor{\bsnm{Nikolakopoulos}, \binits{G.}}:
\batitle{{Dataset collection from a SubT environment}}.
\bjtitle{Robotics and Autonomous Systems}
\bvolume{155},
\bfpage{104168}
(\byear{2022})
\end{barticle}
\endbibitem

%%% 27
\bibitem[\protect\citeauthoryear{Hart et~al.}{1968}]{Hart1968}
\begin{barticle}
\bauthor{\bsnm{Hart}, \binits{P.E.}},
\bauthor{\bsnm{Nilsson}, \binits{N.J.}},
\bauthor{\bsnm{Raphael}, \binits{B.}}:
\batitle{A {Formal} {Basis} for the {Heuristic} {Determination} of {Minimum} {Cost} {Paths}}.
\bjtitle{IEEE Transactions on Systems Science and Cybernetics}
\bvolume{4}(\bissue{2}),
\bfpage{100}--\blpage{107}
(\byear{1968})
\end{barticle}
\endbibitem

%%% 28
\bibitem[\protect\citeauthoryear{Duchoň et~al.}{2014}]{Duchon2014}
\begin{barticle}
\bauthor{\bsnm{Duchoň}, \binits{F.}},
\bauthor{\bsnm{Babinec}, \binits{A.}},
\bauthor{\bsnm{Kajan}, \binits{M.}},
\bauthor{\bsnm{Beňo}, \binits{P.}},
\bauthor{\bsnm{Florek}, \binits{M.}},
\bauthor{\bsnm{Fico}, \binits{T.}},
\bauthor{\bsnm{Juri{\v{s}}ica}, \binits{L.}}:
\batitle{{Path Planning with Modified a Star Algorithm for a Mobile Robot}}.
\bjtitle{Procedia Engineering}
\bvolume{96},
\bfpage{59}--\blpage{69}
(\byear{2014})
\end{barticle}
\endbibitem

%%% 29
\bibitem[\protect\citeauthoryear{Armeni et~al.}{2019}]{Scene_Graph_ICCV}
\begin{bchapter}
\bauthor{\bsnm{Armeni}, \binits{I.}},
\bauthor{\bsnm{He}, \binits{Z.}},
\bauthor{\bsnm{Zamir}, \binits{A.}},
\bauthor{\bsnm{Gwak}, \binits{J.}},
\bauthor{\bsnm{Malik}, \binits{J.}},
\bauthor{\bsnm{Fischer}, \binits{M.}},
\bauthor{\bsnm{Savarese}, \binits{S.}}:
\bctitle{3d scene graph: A structure for unified semantics, 3d space, and camera}.
In: \bbtitle{2019 IEEE/CVF International Conference on Computer Vision (ICCV)},
pp. \bfpage{5663}--\blpage{5672}
(\byear{2019})
\end{bchapter}
\endbibitem

%%% 30
\bibitem[\protect\citeauthoryear{Chang et~al.}{2023}]{Chang2023}
\begin{barticle}
\bauthor{\bsnm{Chang}, \binits{X.}},
\bauthor{\bsnm{Ren}, \binits{P.}},
\bauthor{\bsnm{Xu}, \binits{P.}},
\bauthor{\bsnm{Li}, \binits{Z.}},
\bauthor{\bsnm{Chen}, \binits{X.}},
\bauthor{\bsnm{Hauptmann}, \binits{A.}}:
\batitle{{A Comprehensive Survey of Scene Graphs: Generation and Application}}.
\bjtitle{IEEE Transactions on Pattern Analysis and Machine Intelligence}
\bvolume{45}(\bissue{1}),
\bfpage{1}--\blpage{26}
(\byear{2023})
\end{barticle}
\endbibitem

\end{thebibliography}

\section*{Statements and Declarations}
\subsection*{Funding}
The authors declare that no funds, grants, or other support were received during the preparation of this manuscript.
%Open access funding provided by Lulea University of Technology.
\subsection*{Competing Interests}
The authors have no relevant financial or non-financial interests to disclose.
\subsection*{Author Contributions}
Scott Fredriksson: Development and implementation, main manuscript contributor. Akshit Saradagi and George Nikolakopoulos: Manuscript contributions and advisory role.
\subsection*{Ethics approval}
\subsection*{Consent to participate}
\subsection*{Consent to publish}
%\bibliographystyle{plainnat}
%\bibliography{bliography}

%% else use the following coding to input the bibitems directly in the
%% TeX file.

%\begin{thebibliography}{00}

%\input{bliography.bib}

%\end{thebibliography}
%\printbibliography
\end{document}